\documentclass[journal]{IEEEtran}
\usepackage{amsfonts}
\usepackage{algorithm}
\usepackage{subcaption}
\usepackage{algpseudocode}
\usepackage{array}
\usepackage{adjustbox}

\usepackage{textcomp}
\usepackage{stfloats}
\usepackage[colorlinks,urlcolor=blue,linkcolor=blue,citecolor=blue]{hyperref}
\usepackage{url}
\usepackage{verbatim}
\usepackage{graphicx}
\usepackage{cite}
\usepackage{amsmath}
\usepackage{xcolor}
\usepackage[flushleft]{threeparttable}
\usepackage{multirow}

\usepackage[normalem]{ulem}

\newcommand{\replace}[2]{
\ifthenelse{\equal{#1}{} }{}{\textcolor{olive}{\sout{#1}}}%
\ifthenelse{\equal{#2}{} }{}{ \textcolor{olive}{#2}}%
}

\begin{document}

\title{RoTipBot: Robotic Handling of Thin and Flexible Objects using Rotatable Tactile Sensors}

\author{Jiaqi Jiang$^{*}$, Xuyang Zhang$^{*}$, Daniel Fernandes Gomes, Thanh-Toan Do and Shan Luo 
\thanks{Received 9 November 2024; revised 9 February 2025; accepted 23 April. This work was supported by the EPSRC project ``ViTac: Visual-Tactile Synergy for Handling Flexible Materials" (EP/T033517/2). \textit{(Corresponding author: Shan Luo.)} }
% \thansk{$*$ represents equal contributions.}
\thanks{Jiaqi Jiang is with the School of Aerospace Engineering, Beijing Institute of Technology, Beijing 100081, China, and was with the Department of Engineering, King's College London, London WC2R 2LS, U.K. Email:{\tt\footnotesize 
 jiaqi.jiang@ieee.org}.}
\thanks{Xuyang Zhang, Daniel Fernandes Gomes and Shan Luo are with the Department of Engineering, King's College London, London WC2R 2LS, U.K. Emails: {\tt\footnotesize \{xuyang.zhang, danfergo, shan.luo\}@kcl.ac.uk}.}
\thanks{Thanh-Toan Do is with the Department of Data Science and AI, Monash University, Clayton, VIC 3800, Australia. E-mail: {\tt\footnotesize toan.do@monash.edu}.}%
\thanks{$*$ represents equal contributions.}
% \thanks{Digital Object Identifier (DOI): see top of this page.}
}

\markboth{IEEE Transactions on Robotics,~Vol.~41, April~2025}%
{Shell \MakeLowercase{\textit{et al.}}: A Sample Article Using IEEEtran.cls for IEEE Journals}

\maketitle

\begin{abstract}
This paper introduces RoTipBot, a novel robotic system for handling thin, flexible objects.
Different from previous works that are limited to singulating them using suction cups or soft grippers, RoTipBot can count multiple layers and then grasp them simultaneously in a single grasp closure.  
Specifically, we first develop a vision-based tactile sensor named RoTip that can rotate and sense contact information around its tip. Equipped with two RoTip sensors, RoTipBot rolls and feeds multiple layers of thin, flexible objects into the centre between its fingers, enabling effective grasping. Moreover, we design a tactile-based grasping strategy that uses RoTip's sensing ability to ensure both fingers maintain secure contact with the object while accurately counting the number of fed objects. 
Extensive experiments demonstrate the efficacy of the RoTip sensor and the RoTipBot approach. The results show that RoTipBot not only achieves a higher success rate but also grasps and counts multiple layers simultaneously -- capabilities not possible with previous methods. Furthermore, RoTipBot operates up to three times faster than state-of-the-art methods. 
The success of RoTipBot paves the way for future research in object manipulation using mobilised tactile sensors. 
All the materials used in this paper are available at \url{https://sites.google.com/view/rotipbot}.
\end{abstract}

\section{Introduction}

\IEEEPARstart{T}{hin}, flexible objects are common in our daily lives, e.g., plastic wraps in kitchens, printer papers in offices and delicate lab gloves. Handling these objects is becoming increasingly important for assistive robots, driven by the growing demand for their use in household and everyday settings. For example, assistive robots need to be able to turn pages with precision to help the elderly read and use plastic wraps to store food. Therefore, the development of robotic systems capable of handling these thin, flexible materials with precision is essential for the next generation of assistive robots.

However, thin and flexible objects, characterised by their relatively small thickness compared to their length and width, and their ability to bend or flex without breaking, present two primary challenges for assistive robots. First, their thinness often results in layers that stack over each other, making the states beneath the top layer difficult to discern with a camera, resulting in incomplete and noisy perception~\cite{flipbot,zhu2022challenges}. On the other hand, the tendency of such objects to deform may require dexterity and compliance in robotic grippers~\cite{billard2019trends, zhu2022challenges,teeple2022multi}.

\begin{figure}[t]
   \centering
   \includegraphics[width=1\linewidth]{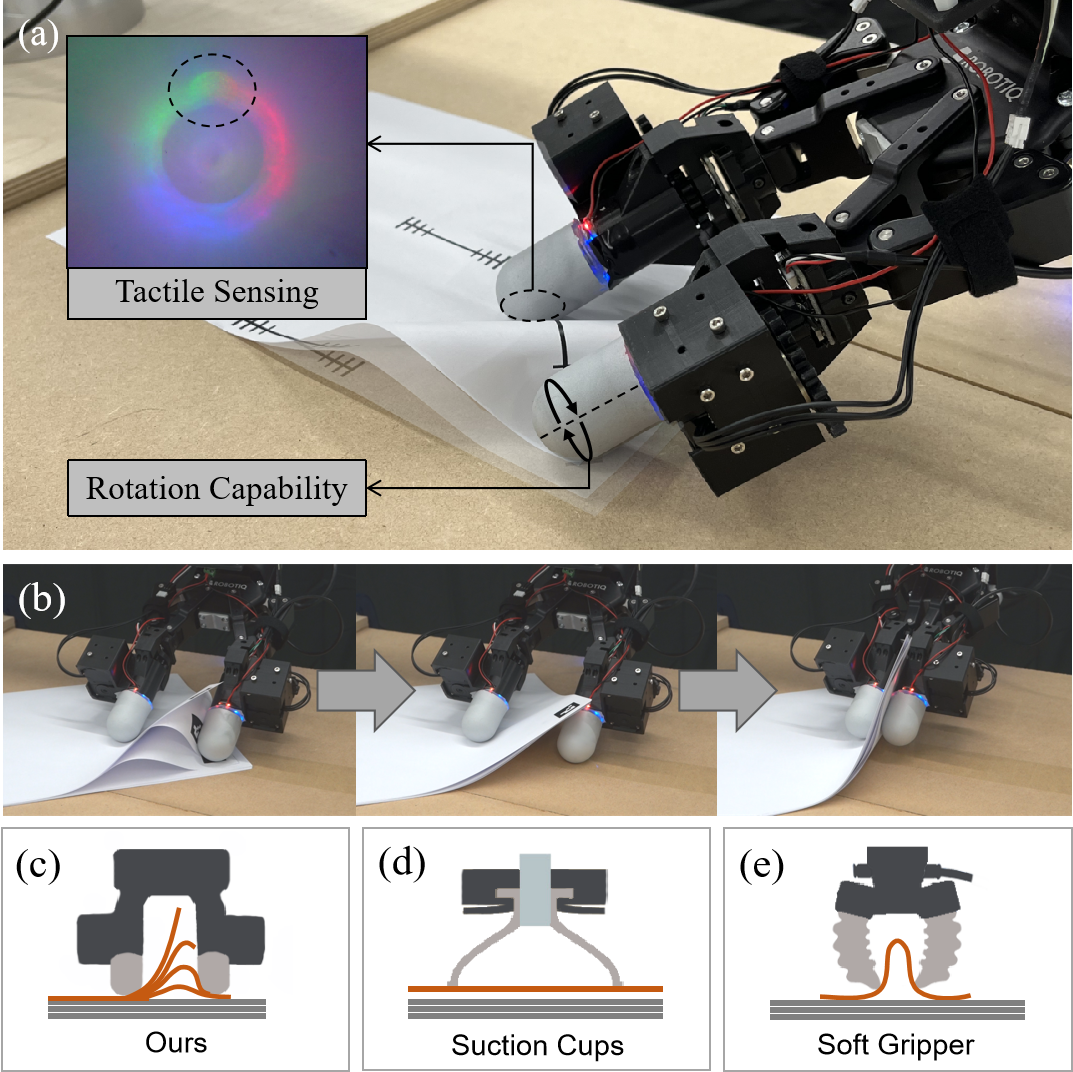}
   \caption{ \textbf{(a)} A demonstration of RoTipBot. The tactile sensors ensure good contact with objects, while the rotation capability feeds multiple layers of thin, flexible objects into the centre for grasping and counting. Different transparencies of the paper represent states at different time steps.
   \textbf{(b)} Snapshots of the feeding process for multiple print papers. 
   \textbf{(c-e)} Sketches comparing RoTipBot to approaches based on suction cups and soft grippers. RoTipBot can count multiple layers and then grasp them simultaneously in a single grasp closure, whereas the other methods cannot. 
   }
\label{fig: 1}
\end{figure}

Existing approaches for grasping thin, flexible objects often bypass these two challenges. Methods typically assume objects are in known positions~\cite{zheng2022autonomous}, or rely on costly force/torque sensors to compensate for the noise in visual perception~\cite{flipbot}, significantly increasing the overall cost. For the second challenge, specialised mechanisms like suction~\cite{koivikko20213d, chin2020multiplexed} and two-finger soft grippers~\cite{jiang2019dynamic, teeple2022multi, flipbot} are used. However, these mechanisms allow only one page in each grasp closure, limiting efficiency. Suction cups struggle with air gaps between sheets, preventing effective vacuum seals, while two-finger soft grippers~\cite{jiang2019dynamic, teeple2022multi, flipbot} lack the dexterity for multi-page grasping where the system first counts multiple layers and then grasps them simultaneously in a single grasp closure.

To address the above challenges in handling thin and flexible objects, in this paper we propose a novel robotic approach named \textit{RoTipBot} that uses rotatable tactile sensors to sense the contacts with these objects and to feed multiple layers of these objects into the centre for grasping and counting.
We develop a new vision-based tactile sensor named \textit{RoTip}, which can sense the entire fingertip area and actively rotate its body. As shown in Fig.~\ref{fig: 1}-(a), RoTip's sensing capability mitigates visual perception noise and ensures good contact with objects. Its rotational capability adds an extra Degree of Freedom, enhancing dexterity in handling thin and flexible objects. A segmentation and projection model is developed to obtain contact areas and surface planes to guide grasping. As shown in Fig.~\ref{fig: 1}-(b), the passive finger of RoTipBot holds the object while the active finger rotates to feed multiple layers into the centre. To adapt to the changes in stacked thickness, the fingers continuously adjust their pose to maintain contact and ensure accurate control when grasping multiple layers. Additionally, tactile sensing is employed for accurate counting of the number of grasped papers. 

We conducted extensive experiments to evaluate the RoTip sensor and RoTipBot approach. The RoTip sensor can accurately estimate contact surface planes with an average error of 1.51$^\circ$.   
The RoTipBot outperforms state-of-the-art methods~\cite{jiang2019dynamic, flipbot} in overall success rate and operates up to 3$\times$ faster. More importantly, it enables the grasping of multiple layers for the first time. 
The success of RoTipBot paves the way for future research in object manipulation using mobilised tactile sensors.

Our contributions can be summarised as follows:
\begin{itemize}
\item We introduce RoTipBot, a novel approach to handling multiple thin, flexible objects, integrating tactile sensing with high dexterity and accuracy.
\item We design RoTip sensors that can sense contacts across the entire fingertip surface with an additional motor enabling active rotation of the sensor surface without angle constraints, significantly enhancing the system’s flexibility and precision in object manipulation.
\item We propose a tactile-based grasping strategy that ensures both fingers maintain secure contact with the object while accurately counting the number of thin, flexible objects being handled.
\item We conducted comprehensive experiments to evaluate RoTipBot, demonstrating its ability to count multiple layers of thin, flexible objects and then grasp them simultaneously in a single grasp closure.
    
\end{itemize}

The rest of this paper is structured as follows. Section~\ref{Section:2} reviews the related works and Section~\ref{Section:3} introduces the design of RoTip and its perception capability; Section~\ref{Section:4} details RoTipBot for thin and flexible object handling; Section~\ref{Section:5} analyses the experimental results; Finally, Section~\ref{Section:6} discusses the work and Section~\ref{Section:7} concludes the paper.

\section{Related Works} \label{Section:2}
\subsection{Vision-based Tactile Sensors}
In recent years, many vision-based tactile sensors have been developed, primarily classified into two main categories: the GelSight family~\cite{yuan2017gelsight}, which processes raw images and employs photometric stereo for shape reconstruction; the TacTip family~\cite{ward2018tactip}, which estimates contact by tracking pin displacement. Unlike TacTip, which is limited by pin resolution, GelSight offers superior resolution and detailed texture information. Our RoTip sensor belongs to the GelSight family.

The first GelSight prototype was developed in 2009 by Johnson and Adelson~\cite{johnson2009retrographic}. Following that, a set of subsequent developments were built with different focuses, such as enhanced lighting setups of GelSight-2017~\cite{yuan2017gelsight} and GelSlim 3.0~\cite{taylor2022gelslim}, sensor reproducibility of DIGIT~\cite{lambeta2020digit}, and a compact shape for robot fingers of~\cite{wang2021gelsight}.
While previous GelSight tactile sensors capture high-resolution tactile images, their limited sensing area poses challenges for complex robotic manipulation. To address this, recent works have introduced finger-shaped tactile sensors, such as GelTip~\cite{gomes2020geltip,gomes2020blocks,gomes2023beyond}, OmniTact~\cite{padmanabha2020omnitact}, Insight~\cite{sun2022soft} and GelSight360~\cite{tippur2023gelsight360}. These sensors can be mounted on off-the-shelf parallel grippers to enhance manipulation tasks with tactile sensing.

\begin{table}
    \centering
    \caption{Comparisons of our RoTipBot against other grippers with rotatable tactile sensors.}
    \label{tab: sensor compare}

    \scalebox{1.0}{
    \begin{tabular}{>{\centering\arraybackslash}p{2cm}|>{\centering\arraybackslash}p{1.2cm}|>{\centering\arraybackslash}p{1.2cm}|>{\centering\arraybackslash}p{2.4cm}}
    \hline
        Feature & TacRot~\cite{tacrot} & TRRG~\cite{yuan2023tactile} & RoTipBot (Ours) \\
        \hline
        \hline
        Unconstrained Rotation & $\times$ & \checkmark & \checkmark \\
        \hline
        Omnidirectional Sensing & $\times$ & $\times$ & \checkmark \\
        \hline
        Perceptible Objects & In-hand Only & In-hand Only & Both In-hand and Out-of-hand \\
        \hline
    \end{tabular}}
\end{table}

However, these setups possess limited degrees of freedom, resulting in a decoupling between tactile sensing and robotic dexterity. To address this limitation, several studies have explored the integration of rotational movement into tactile sensors to enhance functionality, as summarised in Table~\ref{tab: sensor compare}. TouchRoller~\cite{cao2023touchroller} was proposed to enable rapid reconstruction of large contact surfaces through surface rotation, while TacRot~\cite{tacrot} uses a rotatable surface to facilitate in-hand object manipulation. Nonetheless, TacRot~\cite{tacrot} is restricted to sensing only flat elastomer surfaces, and its design with the RGB strip control wire and USB camera cable bundled together limits its rotation range to ±720 degrees. This constraint makes it unsuitable for the continuous rolling motion needed to feed and grasp thin, flexible objects. The tactile-reactive roller grasper (TRRG)~\cite{yuan2023tactile,pan2023hand} further advances this direction by integrating compliant, steerable cylindrical fingertips that allow  unconstrained rotation and perception of the roller's inner side surface. TRRG demonstrates superior performance in in-hand manipulation tasks such as cylindrical object rotation and cable tracing. However, its lack of omnidirectional sensing increases the risk of unintended collisions between the object and the non-sensing area of the grasper, particularly when interacting with objects located outside the hand. As a result, TRRG is not suitable for grasping objects with unknown locations or in scenarios where initial contact cannot be reliably guided by other sensors. In contrast, our proposed RoTip sensor integrates a tactile finger with omnidirectional sensing, enabling reliable detection of contact even under vision localisation errors. Building on this advantage, our RoTipBot employs a tactile-based grasping strategy that does not rely on precise visual alignment.

\subsection{Thin and Flexible Object Grasping}
Deformable object manipulation has been a growing research area with various applications, such as assembling cables in factories~\cite{jin2019robust, she2021cable, pecyna2022visual}, assisted dressing and laundry folding~\cite{zhang2019probabilistic, lee2021learning, sunil2023visuotactile, 2023arXiv230102749Z} and food production~\cite{navarro2017fourier, li2022contact}.
These tasks mainly rely on visual perception~\cite{jin2019robust, zhu2019robotic, lee2021learning, li2022contact, qiu2023robotic}. 
However, the specific challenge of manipulating thin and flexible objects remains relatively unexplored. The small thickness of these objects compared to their length and width and their stacked layers complicate visual perception and grasping.  
% , such as suction~\cite{koivikko20213d, chin2020multiplexed}, soft fingers~\cite{jiang2019dynamic, teeple2022multi, flipbot} and pre-grasp motion~\cite{hang2019pre, babin2019stable}.

Recent approaches to manipulating thin and flexible objects focus on the control design. Several studies achieve thin object grasping via pre-grasp motions such as sliding~\cite{hang2019pre} and scooping~\cite{babin2019stable, zhang2022prying}. In~\cite{hang2019pre}, thin objects are slid to the table edge and grasped from the side. In~\cite{babin2018picking, babin2019stable}, stable grasps of thin objects are achieved through scooping actions. However, scooping methods are generally less precise in separating and flipping exact pages or sheets, often resulting in all layers being grasped together.In~\cite{zhang2022prying}, a prying strategy is proposed to tilt a thin object like a card to create clearance beneath it for a stable pinch grasp. Similarly, in~\cite{ko2020tendon}, an underactuated crawler is used to tilt thin objects for pinch grasps. Zheng et al. \cite{zheng2022autonomous}  simplifies the problem by placing paper on the edge of the table for easy pickup, bypassing the challenge of grasping from the table surface.  

Some other studies focus on specialised grasping mechanisms. 
Suction cups in~\cite{koivikko20213d, chin2020multiplexed} are investigated for handling thin, flexible objects due to their non-damaging, evenly distributed gripping force and simple grasp procedure. Specifically, in~\cite{chin2020multiplexed}, a two-finger gripper integrated with a suction cup is developed for grasping different kinds of objects including thin, flexible ones. However, the effectiveness of the suction mechanism is limited by the need for a smooth surface to create a vacuum seal. When attempting to pick up multiple sheets, air gaps between them prevent the formation of an even vacuum seal, resulting in ineffective or incomplete picks.

Another mechanism used for grasping delicate materials is the soft gripper. As discussed in~\cite{teeple2022multi}, the three-axis compliance of a soft gripper provides unique benefits for grasping thin, flexible objects. In~\cite{jiang2019dynamic}, a two-finger soft gripper is used to deform an object for a pinch grasp, but this approach requires knowledge of physical properties, such as friction, to model object deformation dynamics. Building on this, FlipBot introduced in~\cite{flipbot} uses a two-finger soft gripper with a force/torque sensor to singulate and grasp thin and flexible objects by integrating exteroceptive and proprioceptive perceptions. However, it uses a force/torque sensor that only provides contact information at a single point, and can only grasp one page at a time, limiting its efficiency. 

In contrast, our RoTipBot offers a tactile-based grasping strategy that ensures both fingers maintain secure contact with the object while accurately counting the number of thin, flexible objects. In this way, multiple thin, flexible objects can be grasped simultaneously in a single grasp closure. 
Although there are several studies~\cite{agboh2022multi, li2024grasp} on grasping multiple rigid objects, the problem of grasping multiple thin and flexible objects remains largely unexplored in current literature.

\begin{figure*}[t]
\centering
   \includegraphics[width=\linewidth]{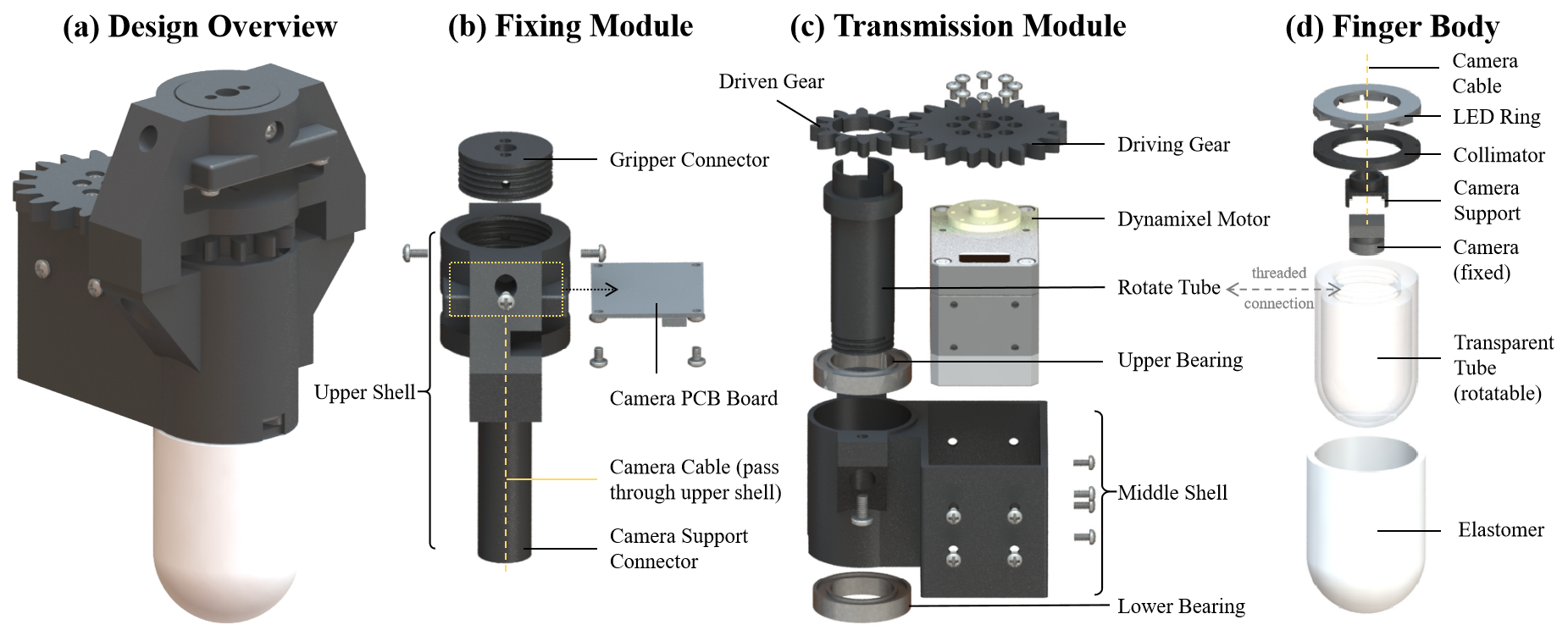}
   \caption{\textbf{From Left to Right:} \textbf{(a) Design overview} of the RoTip sensor, and the exploded view of RoTip's three modules: \textbf{(b) fixing module} that serves as the structural backbone, ensuring stability and interconnection among sensor components; \textbf{(c) transmission module} that facilitates rotational movement, enabling dynamic functionality; and \textbf{(d) finger body} that constitutes the tactile interface, providing the tactile sensing capabilities.}
\label{fig: 2}
\end{figure*}

\section{Hardware Setup} \label{Section:3}

In this section, we introduce RoTip, a vision-based tactile sensor that can sense the entire fingertip area and rotate along the vertical axis, as shown in Fig.~\ref{fig: 2}. We then present the hardware setup for handling thin and flexible objects, including the RoTip sensor.

\subsection{The RoTip Sensor}
The proposed RoTip sensor comprises three primary modules: the fixing module, the transmission module and the finger body, as shown in Fig.~\ref{fig: 2}-(b), (c) and (d), respectively. The fixing module is used for secure attachment and precise positioning of the transmission module and the finger body, ensuring stability during operation. Meanwhile, the transmission module is used to drive the rotational motion, enabling dynamic functionality. Lastly, the finger body serves as the tactile interface, providing the sensor's tactile perception capabilities. The fabrication details can be found on our website.

The specifications of the designed sensor are detailed as follows. Its payload capacity can be evaluated from two perspectives: the grasping payload, and the rolling mechanism payload. For grasping, the sensor can handle up to 3.4 kg, derived from the maximum payload of the UR5e robotic arm minus the weight of the gripper. For rolling, the RoTip sensor achieves a max payload of 20.3N when the actuator operates at a speed of 90°/s, which is sufficient for handling thin, flexible objects, such as the print paper that requires 2N of force to roll. To provide a clear visual reference, the dimensions of the RoTip sensor have been included in Fig.~\ref{fig: 2}. To reduce the processing load for tactile images, the sampling frequency of the RoTip sensor has been set to 30 Hz with a resolution of 640$\times$480 pixels. 

\begin{figure}
    \centering
    \includegraphics[width=1\linewidth]{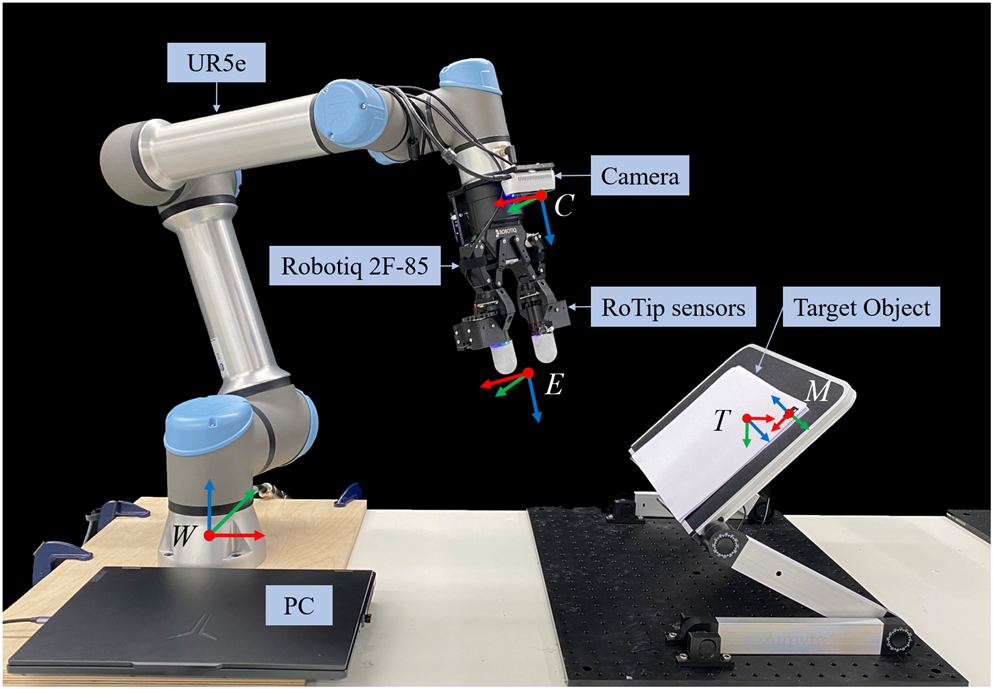}
    \caption{ \textbf{ Hardware Setup and Coordinates}. The hardware setup consists of a 6-DOF UR5e robotic arm, a Robotiq 2F-85 gripper, an Intel RealSense D435 RGB-D camera, and a laptop stand for holding the printer papers or other thin and flexible objects. The fingertips of the Robotiq gripper are replaced with two RoTip sensors. $W$ and $E$ represent the world coordinate that is located at the base of UR5e and the end-effector coordinate, respectively. $C$ and $M$ represent the camera coordinate and the marker coordinate, respectively. $T$ is the target coordinate related to the marker coordinate and used to direct the movement of the end-effector. }
    \label{fig: 3}
\end{figure}

\subsection{Robotic Setup}
As shown in Fig. \ref{fig: 3}, our robot setup consists of a 6-DOF UR5e robot\footnote{https://www.universal-robots.com/products/ur5-robot/}, a Robotiq 2F-85 adaptive gripper\footnote{https://robotiq.com/products/2f85-140-adaptive-robot-gripper}, and an Intel RealSense D435 RGB-D camera\footnote{https://www.intelrealsense.com/depth-camera-d435/}. The fingertips of the Robotiq gripper are replaced with two RoTip sensors. All the above devices are connected to a laptop equipped with an Nvidia RTX 4060 GPU. Furthermore, for evaluating the grasp in various poses, we utilise an adjustable stand mounted to the test board on the table. This stand is designed to hold the thin and flexible object intended for grasping, enabling consistent and reliable experimentation.

In this work, we establish five distinct coordinate systems, using the base of the UR5e robotic arm as the origin for the world coordinate system, denoted as $\boldsymbol{W}$. The end-effector coordinate, denoted as $\boldsymbol{E}$, is initially positioned at the centre of the two RoTip sensors and can be dynamically tracked via the UR5e ROS driver. The camera coordinate system, $\boldsymbol{C}$, is located at the RGB component of the RealSense camera. To streamline the calibration process between the camera and the end-effector, a specialised mounting component is designed, ensuring only translational offsets exist between these two coordinates. Additionally, we identify the corner of the thin and flexible object, marked by an ArUco marker, as the marker coordinate system, represented as $\boldsymbol{M}$. The final target coordinate for the grasping action, denoted as $\boldsymbol{T}$, is established to direct the movement of the end-effector. The transformation $\boldsymbol{\mathrm{T}_M^T}$ from $\boldsymbol{T}$ to $\boldsymbol{M}$ is predefined to make the line connecting the two fingers parallel to the angle bisector of the object's corner.

\section{Thin and Flexible Object Grasping} \label{Section:4}
% find a name for our approach
In this section, we introduce RoTipBot, an approach that uses the RoTip rotatable tactile sensors to count multiple layers and then grasp them simultaneously in a single grasp closure. RoTipBot uses a two-finger gripper to handle thin and flexible objects: one \textbf{passive} finger holds the object, while the other \textbf{active} finger rotates to feed multiple sheets into the centre between the two fingers, allowing them to be grasped all at once. This method prevents pinching the paper from the same side, thereby avoiding potential damage and ensuring a more secure grasp. Though in this work we demonstrate the RoTipBot with a two-finger gripper, it is generalisable to a robotic hand with more than two fingers. 

First, we use an RGB-D camera to obtain the rough grasping pose for handling thin and flexible objects.
Following that, we propose a tactile-based adjustment method based on RoTip’s perception capability to adjust the pose of the end-effector, ensuring grasp stability. Finally, we propose a tactile-based paper counting approach and a continuous adjustment method to address the changes in stacked thickness during the feeding process of multiple layers of the object for grasping.

\subsection{Vision-Guided Grasp Pose Estimation}\label{Sec: grasp generation}
In this subsection, we estimate the real-world grasping poses for squeezing and feeding thin and flexible objects with an RGB-D camera. To facilitate corner detection, we employ ArUco markers, streamlining the process and allowing us to concentrate on the primary challenges addressed by the proposed RoTip sensors: overcoming incomplete visual perception of stacked layers and enhancing dexterity in handling thin and flexible objects.

\begin{figure*}
    \centering
    \includegraphics[width=0.9\linewidth]{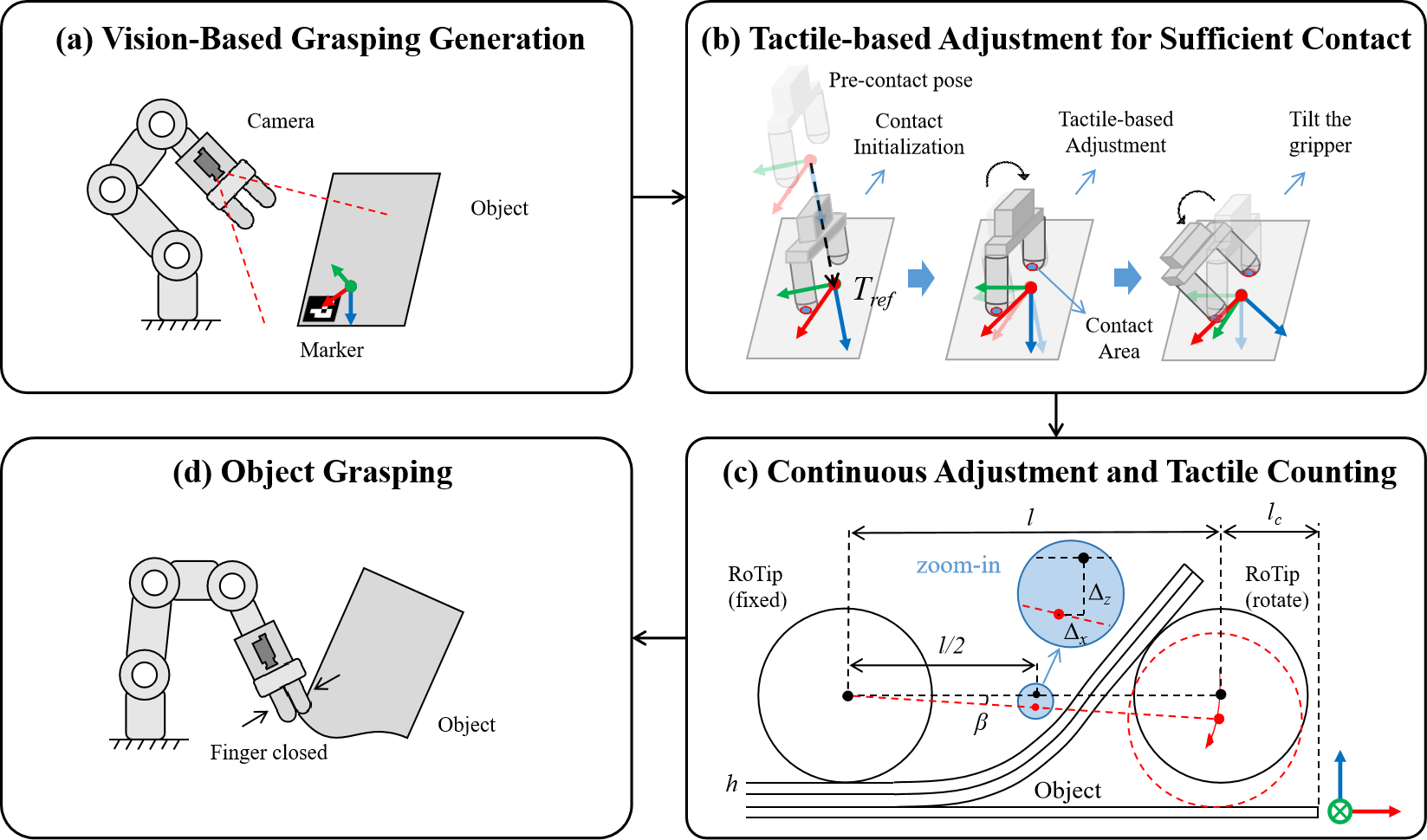}
    \caption{An overview of our RoTipBot for thin and flexible handling. \textbf{(a) Vision-Based Grasping Generation:} Given an RGB-D image obtained from a camera, we generate the grasp proposal for guiding the robot to contact and grasp the object. \textbf{(b) Tactile-based Adjustment for Two-finger Sufficient Contact:} To compensate for the noise from visual perception, RoTip’s sensing capability is then utilised to adjust the robot’s end-effector. Once both RoTip sensors are in contact with the object, the end-effector will be rotated around its $x$-axis to an angle inclined to the object for feeding and grasping. \textbf{(c) Continuous Adjustment and Tactile Counting:} A continuous pose adjustment approach is proposed to ensure two-finger contact while feeding multiple thin and flexible objects. Tactile sensing is also employed to count the number of fed pages, a process referred to as \textit{tactile counting}. \textbf{(d) Object Grasping: } Finally, the gripper is closed to pick up the objects. }
    \label{fig: 4}
\end{figure*}

To simplify the detection of corners on various thin and flexible objects, we attach an ArUco marker to the corner as shown in Fig.~\ref{fig: 4}-(a). The rationale for selecting the corners as the optimal locations for handling thin and flexible objects is detailed in Appendix~\ref{Section:grasp}.
With the known side length of the ArUco marker, we can estimate the transformation matrix from the camera's coordinate system to the marker's coordinate system, denoted as $\boldsymbol{\mathrm{T}_C^M}$. This is achieved by minimising reprojection error through non-linear optimisation. To account for estimation errors that can occur with ArUco markers in real-world applications, we use depth information to adjust the estimated $\boldsymbol{\mathrm{T}_C^M}$. 
First, we crop the depth map around the corners, similar to the approach in~\cite{flipbot}, to fit the object plane. Then, we adjust the rotation matrix of $\boldsymbol{\mathrm{T}_C^M}$ to be parallel to the depth-fitted plane, and we update the translation vector of $\boldsymbol{\mathrm{T}_C^M}$ based on the depth of the ArUco marker centre. Then, the target pose in the world frame represented with the transformation matrix $\boldsymbol{\mathrm{T}_W^T}$ can be calculated as:
\begin{equation} \label{eq: 7}
    \boldsymbol{\mathrm{T}_W^T} = \boldsymbol{\mathrm{T}_W^E}*\boldsymbol{\mathrm{T}_E^C}*\boldsymbol{\mathrm{T}_C^M}*\boldsymbol{\mathrm{T}_M^T},
\end{equation}
where $\boldsymbol{\mathrm{T}^E_W}$ represents the transformation matrix of the end-effector in the world coordinate;
$\boldsymbol{\mathrm{T}_E^C}$ and $\boldsymbol{\mathrm{T}_M^T}$ represent the transformation matrix from the end-effector coordinate system to the camera coordinate system and from the marker coordinate system to the target coordinate system, respectively.

\subsection{Tactile-based Adjustment for Sufficient Contact}
Given an estimated target pose from Sec.~\ref{Sec: grasp generation}, moving directly to the target pose may result in collisions or insufficient contact due to the inherent positioning uncertainty of the vision system. Therefore, in this subsection, we introduce a tactile-based adjustment method that ensures both fingers securely contact thin, flexible objects to guarantee a secure grasp, i.e., two-finger contact.  As is shown in Fig.~\ref{fig: 4}-(b), the tactile-based adjustment includes two key steps, i.e., contact initialisation and pose adjustment.

\subsubsection{Contact Initialisation} \label{Section: Feedback Control}
The end-effector is moved to a pre-contact pose, positioned with a translational offset along the $z$-axis relative to the target pose. A normal contact motion, using feedback control guides the robot towards the object, and terminates when a RoTip finger makes initial contact with the object.

% For the feedback control, we define the pose error $\boldsymbol{\mathrm{E}}_{\boldsymbol{\mathrm{T}}}$ as the $SE(3)$ transformation that moves the observed pose $\boldsymbol{\mathrm{T}_\mathrm{W}^{\mathrm{O}}}$ to the target pose $\boldsymbol{\mathrm{T}_{\mathrm{W}}^{\mathrm{T}}}$ in the same coordinate frame:
% %  as~\cite{lloyd2023pose}
% \begin{equation}
% 	\label{eqn:local_coord_frame_error}
% 	\boldsymbol{\mathrm{E}}_{\boldsymbol{\mathrm{T}}}
%     \,=\, (
% 	\boldsymbol{\mathrm{T}_\mathrm{W}^{\mathrm{O}}})^{-1}\boldsymbol{\mathrm{T}_{\mathrm{W}}^{\mathrm{T}}}.
% \end{equation}
% % where $\boldsymbol{\mathrm{X}}$ and $\boldsymbol{\mathrm{X}}_{\mathrm{ref}}$ are specified in the world frame located at the base of the robot. 

% To define the control operations, we project this pose error into the exponential coordinates for the Lie algebra $\mathfrak{se}(3)$, mapped onto the vector tangent space $\mathbb{R}^6$:
% \begin{equation}
% 	\label{eqn:local_tangent_space_error}
% 	\boldsymbol{\mathrm{e}}_{\boldsymbol{\mathrm{T}}}%(\boldsymbol{\mathrm{X}})
% 	\,=\,
% 	\ln(\boldsymbol{\mathrm{E}}_{\boldsymbol{\mathrm{T}}})^{\vee}.
% \end{equation}

We first design the pose error as $\boldsymbol{e_{\mathrm{T}}} = \ln(\boldsymbol{\mathrm{T}}^{-1} \boldsymbol{\mathrm{T}}_{\mathrm{ref}})^{\vee}$ between the observed pose $\boldsymbol{\mathrm{T}}$ to the target pose $\boldsymbol{\mathrm{T}}_{\mathrm{ref}}$. There are two steps in establishing the contact. 
When the gripper is far from the object, the Euclidean normal of the translation component of $\boldsymbol{e_{\mathrm{T}}}$'s is larger than a predefined threshold $\epsilon$ (set as 0.02 in our case), i.e., $\lVert \boldsymbol{e_{\mathrm{T}}}[0:3]\rVert > \epsilon$, a PD controller is used to control the robot to approach the surface of the thin and flexible object, i.e., $\boldsymbol{\mathrm{u}}(t) = \boldsymbol{\mathrm{K}}_{p} \boldsymbol{\mathrm{e}}_{\boldsymbol{\mathrm{T}}}(t) + \boldsymbol{\mathrm{K}}_{d} \, \frac{d \boldsymbol{\mathrm{e}_{\boldsymbol{\mathrm{T}}}}}{dt}(t)$, where $\boldsymbol{\mathrm{K}}_{p}$ and $\boldsymbol{\mathrm{K}}_{d}$ are $6 \times 6$ diagonal gain matrices associated with the error and derivative errors at time $t$.
Once the gripper is close enough to the object, i.e., $\lVert \boldsymbol{e_{\mathrm{T}}}[0:3]\rVert \leq \epsilon$, a feedforward control $\boldsymbol{\mathrm{v}}(t)$ is taken for $\boldsymbol{\mathrm{u}}(t)$ to move the end-effector towards the grasp pose to simplify the control.

\begin{algorithm}[t] 
\caption{Tactile-based Adjustment for Two-Finger Contact} \label{alg::1}
\begin{algorithmic}[1]
\State Move to a pre-contact pose based on the target pose $\boldsymbol{\mathrm{T}}_{\mathrm{ref}}$  \;
\State Number of sufficient contact fingers: $N = 0$ \;
\State Pose Error: $ \boldsymbol{\mathrm{e}}_{\boldsymbol{\mathrm{T}}} = \ln(\boldsymbol{\mathrm{T}}^{-1} \boldsymbol{\mathrm{T}}_{\mathrm{ref}})^{\vee}.$ \;
\While {True}
\If {$N==0$} \quad \textit{Sec.~\ref{Section: Feedback Control} Contact Initialisation}
    \If {$\lVert \boldsymbol{e_{\mathrm{T}}}[0:3]\rVert > \epsilon$}
    \State $\boldsymbol{\mathrm{u}}(t)
	=  \boldsymbol{\mathrm{K}}_{p} \, \boldsymbol{\mathrm{e}}_{\boldsymbol{\mathrm{T}}}(t)
	+ \boldsymbol{\mathrm{K}}_{d} \, \frac{d \boldsymbol{\mathrm{e}_{\boldsymbol{\mathrm{T}}}}}{dt}(t),$
    \Else
    \State $\boldsymbol{\mathrm{u}}(t) = \boldsymbol{\mathrm{v}}(t) $
   
    \EndIf
\ElsIf {$N==1$}   \quad \textit{Sec.~\ref{Section: Tactile-based Adjustment} Pose Adjustment}
    \State $\boldsymbol{\mathrm{T}}_{\mathrm{ref}} = \boldsymbol{\mathrm{T}}_{\mathrm{ref}} * \boldsymbol{\mathrm{\left[R, \boldsymbol{\mathrm{0}}\right]}}$ 
    \State Move to the refined pre-contact pose
    \State N = 0
% \Else  
\ElsIf {$N==2$}
    \State Break
\EndIf
\State update $\boldsymbol{e_\mathrm{T}}$ and $N$

\EndWhile
\end{algorithmic}
\end{algorithm}

\begin{figure*}
    \centering
    \includegraphics[width=1\linewidth]{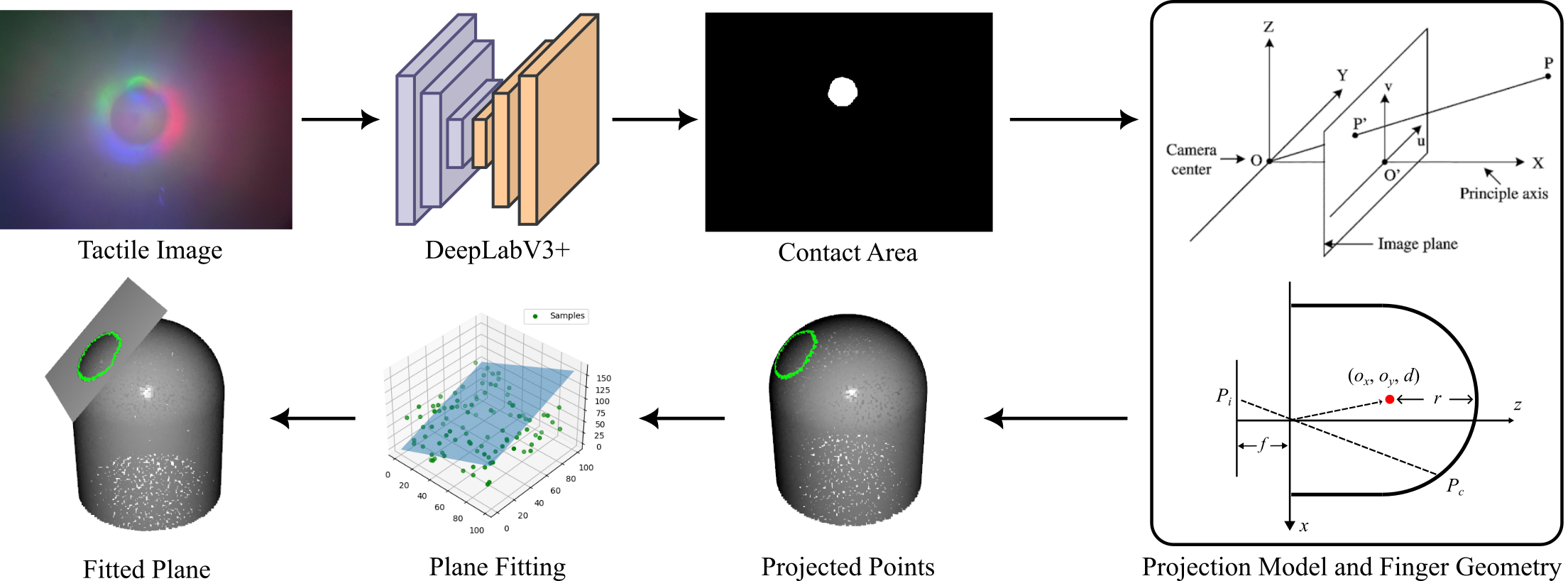}
    \caption{\textbf{Pipeline for tactile-based thin object normal prediction.} Given a tactile image as input, a fine-tuned DeepLab v3+~\cite{chen2018encoder} network is used to segment the contact area of the RoTip sensor. Then, the contour of the contact area can be obtained based on the camera projection model and the finger geometry. Finally, a RANSAC-based plane fitting approach is used to derive the parameters of the contact plane. }
    \label{fig: 5}
\end{figure*}

\subsubsection{Pose Adjustment for Two-Finger Contact} \label{Section: Tactile-based Adjustment} To ensure both RoTip sensors make sufficient contact with the object and achieve two-finger contact, the end-effector pose must be refined based on the tactile feedback. Specifically, achieving two-finger contact involves aligning the end-effector $z$-axis normal vector \( \mathbf{n}_1 \) with the contact plane normal vector \( \mathbf{n}_2 \). 

This alignment can be achieved by calculating the rotation matrix $\boldsymbol{\mathrm{R}}$ to adjust the end-effector pose using Rodrigues' formula as follows:

\begin{equation}
\mathbf{R} = \mathbf{I} + (\sin \theta) \mathbf{A} + (1 - \cos \theta) \mathbf{A}^2
\end{equation}
where $\theta = \arccos(\mathbf{n}_1 \cdot \mathbf{n}_2)$ is the rotation angle and \( \mathbf{A} \) is the skew-symmetric matrix of the rotation axis \( \mathbf{a} = [a_x, a_y, a_z] = \mathbf{n}_1 \times \mathbf{n}_2 \):
\begin{equation}
\mathbf{A} = \begin{pmatrix} 0 & -a_z & a_y \\ a_z & 0 & -a_x \\ -a_y & a_x & 0 \end{pmatrix} .
\end{equation}

To obtain the contact plane normal vector \( \mathbf{n}_2 \), the captured tactile image is first fed to a fine-tuned DeepLab v3+ network~\cite{chen2018encoder} to segment the contact area as shown in Fig.~\ref{fig: 5}. Following this, the camera's projection model is used to reconstruct the point cloud representing the boundary of the contact area. Given the estimated boundary of contact areas, a RANSAC-based plane fitting method is used to estimate the surface normal of the contact plane in 3D space. By identifying the contact plane, the robot could adjust its end-effector pose for optimal contact.

\subsubsection{The Calculation of Contact Points}
Based on the general camera projection model~\cite{szeliski2022computer}, the segmented contact areas in pixel coordinates, represented in homogeneous form as $(u,v,1)$, can be converted to normalised image coordinates $(x_i, y_i, 1)$ for the $i^{th}$ pixel in the image frame as follows:

\begin{equation}
\left[\begin{array}{l}
x_i \\
y_i \\
1
\end{array}\right]=\left[\begin{array}{lll}
\frac{1}{f_x} & 0 & -\frac{c_x}{f_x} \\
0 & \frac{1}{f_y} & -\frac{c_y}{f_y} \\
0 & 0 & 1
\end{array}\right] \cdot\left[\begin{array}{l}
u \\
v \\
1
\end{array}\right],  
\end{equation}

\noindent in which $f_x$ and $f_y$ are focal lengths in $x$ and $y$, respectively; $c_x$ and $c_y$ are the offsets in the image frame.

The normalised image position $(x_i, y_i, 1)$ also represents the direction of the light ray travelling through the sensor membrane, where the contact point $P_c=(x,y,z)$ on the sensor surface must lie on. Hence, we have
\begin{equation}
\label{eq3}
[x, y, z] = k*[x_i, y_i, 1] \\ 
, 
\end{equation}
\noindent where $k \in \mathbb{R}$ is a scale parameter. 

As illustrated in the right subfigure of Fig.~\ref{fig: 5}, the sensor surface can be modelled as a combination of a hemisphere and an opened cylinder, both sharing the same radius $r$. The central axis of the cylinder surface is parallel to the $z$-axis. Therefore, the centre point of the hemisphere can be set to $(o_x,o_y,o_z)$, where $o_z$ is the distance from the centre point of the base of the hemisphere to the centre point of the base of the sensor. Then, the location of any point on the sensor surface $P_c$ can be represented as follows: 
\begin{equation}
\label{eq4}
\begin{cases}(x-o_x)^2+(y-o_y)^2+(z-o_z)^2=r^2 & \text { for } z>o_z \\ (x-o_x)^2+(y-o_y)^2=r^2 & \text { for } z \leq o_z\end{cases}
\end{equation}

By solving Eq.~\ref{eq3} and Eq.~\ref{eq4}, we can obtain the contact location on the sensor surface.

\begin{figure}
    \centering
    \includegraphics[width=1\linewidth]{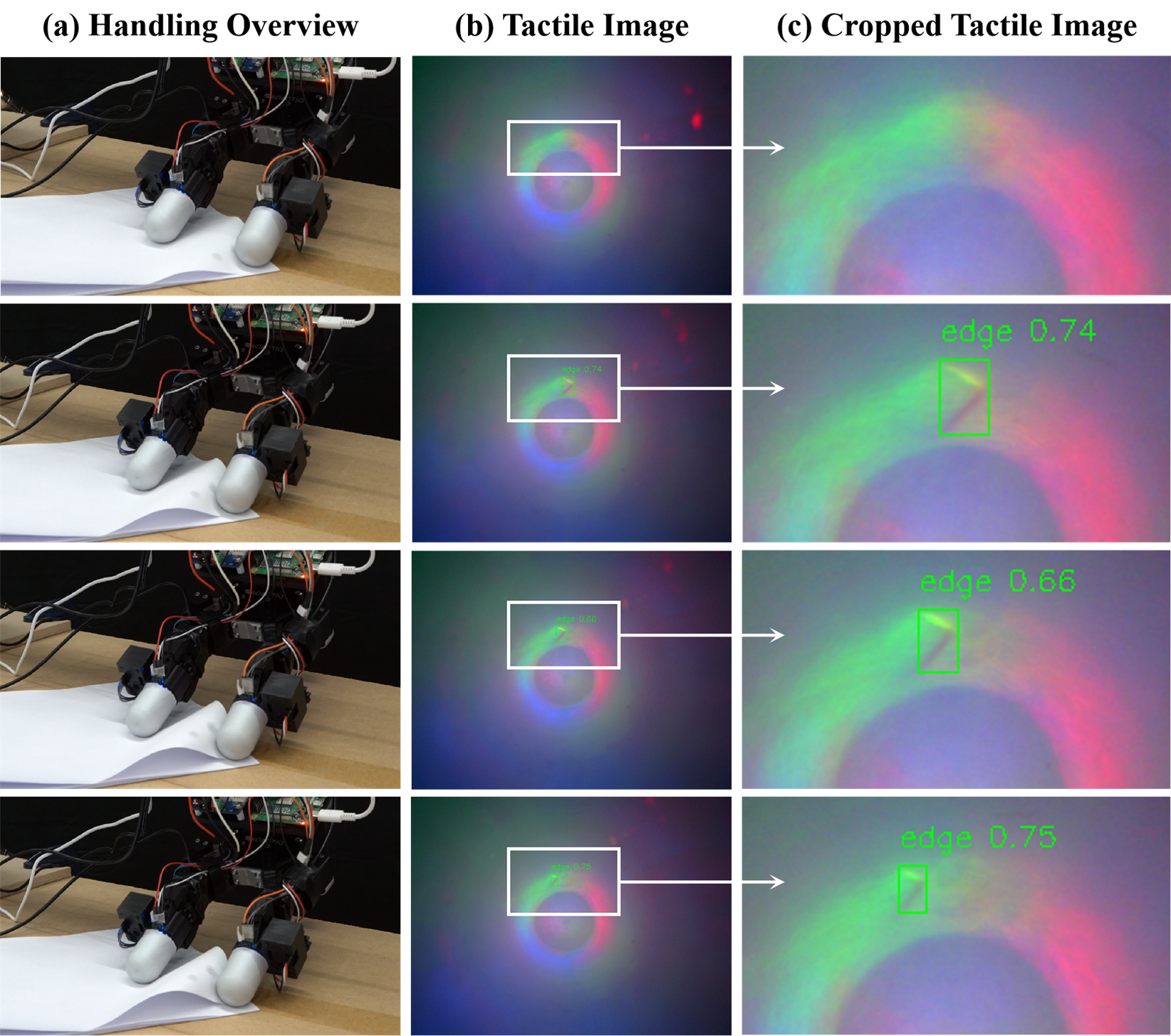}
    \caption{Demonstration of page counting with tactile feedback. (a) The vision snapshots during feeding the stacked papers; (b) The paper edges can be captured in the tactile images to count the number of fed papers. (c) The zoomed-in contact areas in tactile images. }
    \label{fig: 6}
\end{figure}

\subsection{Continuous Adjustment and Tactile Counting} \label{Section: Continuous Adjustment} 
Although the initial pose adjustment in Sec.~\ref{Section: Tactile-based Adjustment} could secure two-finger contact at the beginning, continuous adjustment of the end-effector is necessary to address the changes in the stacked thickness during the feeding of multiple sheets. As shown in Fig.~\ref{fig: 4}-(c), a 3-dimensional adjustment $(\beta, \Delta_x, \Delta_z)$ is required to refine the end-effector pose: 
\begin{equation}
  \begin{cases}
    \beta & = \arcsin{\frac{n*h}{l}} \\
    \Delta_x & = -l/2*(1-\cos{\beta}) \\
    \Delta_z & = -l/2*\sin{\beta}
  \end{cases}
\end{equation}
where $\beta$ is the rotational adjustment along the $y$-axis, and $\Delta_x$ and $\Delta_z$ are the translational adjustments along the $x$-axis and $z$-axis of the local coordinate, respectively. The axes of the local coordinate are represented with Red, Green and Blue arrows at the right corner of Fig.~\ref{fig: 4}-(c). Here, $n$, $h$ and $l$ represent the number of fed layers, the thickness of each layer, and the distance between two RoTip sensors, respectively. This adjustment must then be transformed from the current local coordinate system to the world coordinate system.

Additionally, tactile sensing is employed to count the number of fed pages, a process referred to as \textit{tactile counting}, as shown in Fig.~\ref{fig: 6}. A YOLOv11 model is used to detect the paper boundaries including edges and corners during the feeding process of multiple thin and flexible sheets. The detection focuses on the edges of the paper to track movement. When the centre of a detected edge crosses a predefined threshold, the paper is marked as ``held" for rolling by the RoTipBot. The count increases by 1 when a new edge appears, indicating a new paper is being rolled by the sensor.

\section{Experimental Results} \label{Section:5}
In this section, we will conduct comprehensive experiments to evaluate our proposed RoTipBot approach for handling thin and flexible objects. The experiments will focus on four key aspects: (1) identifying the optimal grasping locations for handling these objects; (2) exploring the significance of tactile-based adjustment to ensure consistent two-finger contact; (3) assessing the effectiveness of the continuous adjustment and tactile counting with different kinds of thin and flexible objects against the state-of-the-art approaches; and (4) exploring additional applications with our RoTip sensors.

\subsection{Experiments on Grasp Position Analysis}
In this subsection, we conduct two experiments to verify that the corners of thin and flexible objects are the optimal locations for squeezing and feeding. Specifically, we perform experiments to compare the force applied to the object during squeezing, and to determine the total number of thin and flexible objects (using papers as an example) grasped at various locations. 

\begin{figure}
    \centering
    \includegraphics[width=\linewidth]{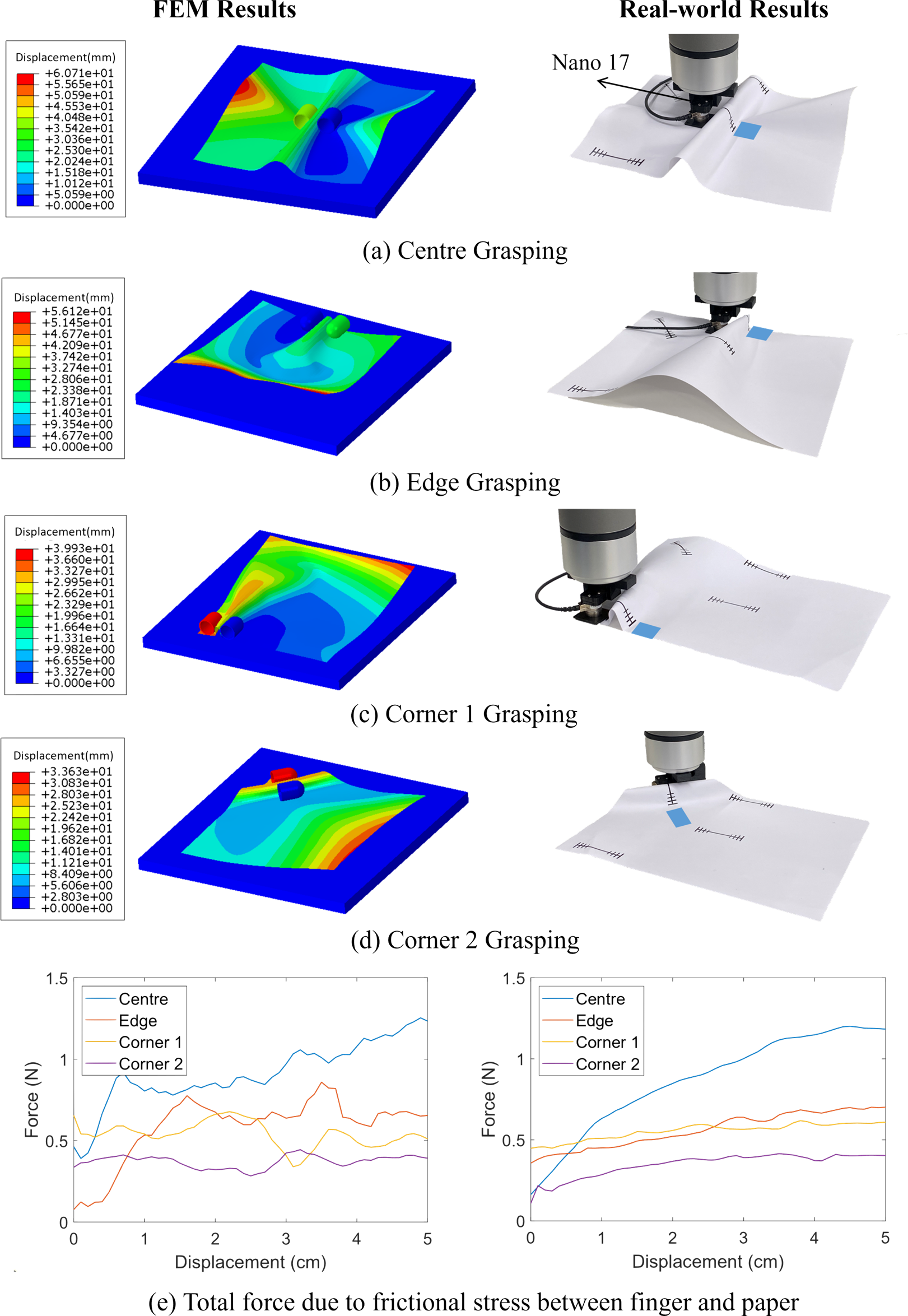}
    \caption{\textbf{Comparison of required forces when squeezing different locations.} The left and right columns show the FEM results and real-world results for estimating the elastic force of the object, respectively. The first four rows show experiments at different locations (i.e., centre, edge, and corners). The last row presents the force plots for squeezing the paper. The results indicate that squeezing along the angle bisector of the object's corner (d) results in the smallest elastic force, thereby requiring the least squeezing force.}
    \label{fig: 7}
\end{figure}

% \subsubsection{Comparison of Required Forces}
To compare the forces during the squeezing of thin and flexible objects, we employ both Finite Element Method (FEM) in simulation and a force sensor in real-world experiments, as shown in Fig.~\ref{fig: 7}.  We performed the squeezing action at various locations on the object, i.e., the centre, edge, and corners. Squeezing the object at its corner is divided into two distinct scenarios, each defined by the direction of applied force. In Case 1, the squeezing is executed parallel to the object's edge, as shown in Fig.~\ref{fig: 7}-(c). Case 2 involves squeezing along the angle bisector of the object corner, as depicted in Fig.~\ref{fig: 7}-(d). 

In the simulation test, FEM is used to simulate the interaction between the RoTip fingers and the object at different locations. The process involves two steps: first, two fingers press the object; then, one of the fingers squeezes the paper by sliding. The simulation provides precise force estimations exerted by the object and visualises the displacement distribution. Details of the simulation configuration are provided in Appendix~\ref{appendix:b}. 

We also conducted a real-world experiment. A Nano 17 force/torque sensor was employed to measure the forces exerted while squeezing the object. This sensor was attached to the end-effector of the UR5e robotic arm, as shown in Fig.~\ref{fig: 7} (right column), where a blue quadrangular area was secured to the table using double-sided tape to simulate the pressing of the passive finger. Although the RoTip sensor has the potential to measure applied force, it does not match the precision of the Nano 17, which is commonly used in previous research~\cite{yuan2017gelsight, sun2022soft, azulay2023allsight} as a benchmark for ground truth.

As shown on the left of Fig.~\ref{fig: 7}-(e), the maximum forces applied by the object show a clear pattern: the centre is the hardest to squeeze, followed by the edges, and finally the corners. Furthermore, squeezing along the angle bisector of the object corner results in the smallest maximum force. 
In our real-world experiments, the same trend is observed, with the corner cases involving smaller squeezing forces. Specifically, the average maximum force applied by the object at the corner (Case 2) was measured at 0.419N, which is notably lower than the forces measured at the corner (Case 1) (0.731N), edge (0.753N) and centre (1.246N). These findings validate the conclusions drawn from our theoretical analysis in Appendix~\ref{Section:grasp}. Moreover, both the FEM visualisation of displacement distribution and real-world snapshots indicate that when squeezing at the corners, the object moves less vigorously around the fingertip, reducing the likelihood of damaging the object.

\begin{figure}
    \centering
       % \fbox{\rule{0pt}{4in} \rule{0.9\linewidth}{0pt}}
       \includegraphics[width=\linewidth]{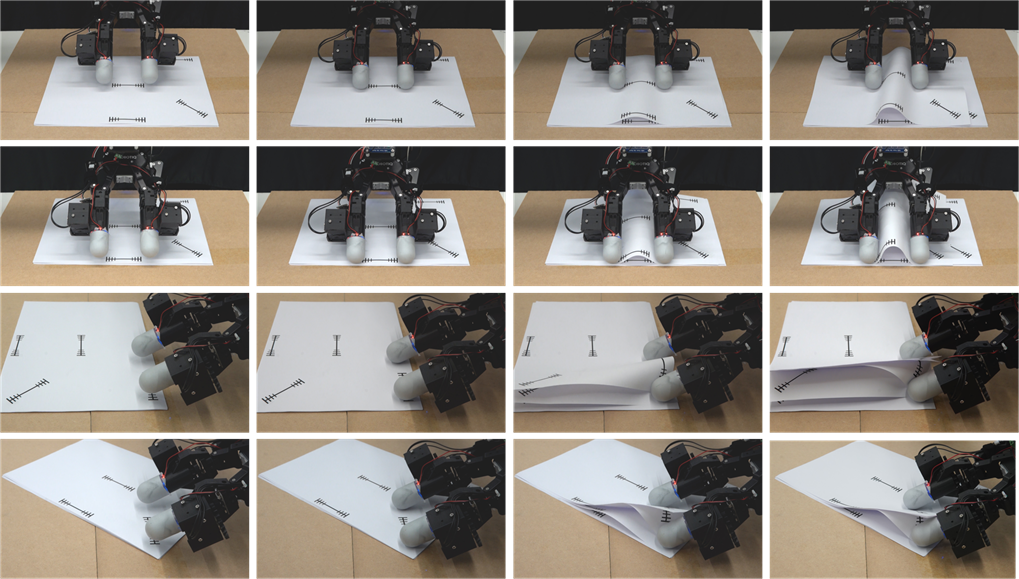}
       \caption{ Snapshots of squeezing at different locations. \textbf{From top to bottom:} the centre, the edge, and the corner (two cases). During the 5-second squeezing, only one page can be squeezed at the centre and edge, while multiple pages can be squeezed at the corners. When squeezing along the angle bisector of the object corner (Case 2), the deformation of the printer paper is less intense. }
       \label{fig: 8}

\end{figure}

To verify that grasping at the corners is the most effective, we conducted grasping experiments in a real-world setting. Our evaluation of performance relied on the key metric of the total number of grasped papers. We maintained a constant actuator speed, set at 90 degrees per second for 5 seconds. After that, the gripper is closed to pick up the object. We captured snapshots of the squeezing process at different locations, shown in Fig.~\ref{fig: 8}. When squeezing along the angle bisector of the object corner (Case 2), the deformation of the printer paper is less intense and squeezing objects around corners leads to the largest number of grasped papers.

\subsection{Tactile-based Adjustment for Two-finger Contact}
In this subsection, we conduct a series of experiments to evaluate the tactile-based adjustment in achieving two-finger contact, a crucial requirement for manipulating the objects. We focus on two key aspects: the accuracy of plane fitting using different modalities, i.e., vision, force sensing and tactile feedback; and the overall success rate of contact between RoTip sensors and the thin and flexible object.

\begin{figure*}
    \centering
    \includegraphics[width = 0.9\linewidth]{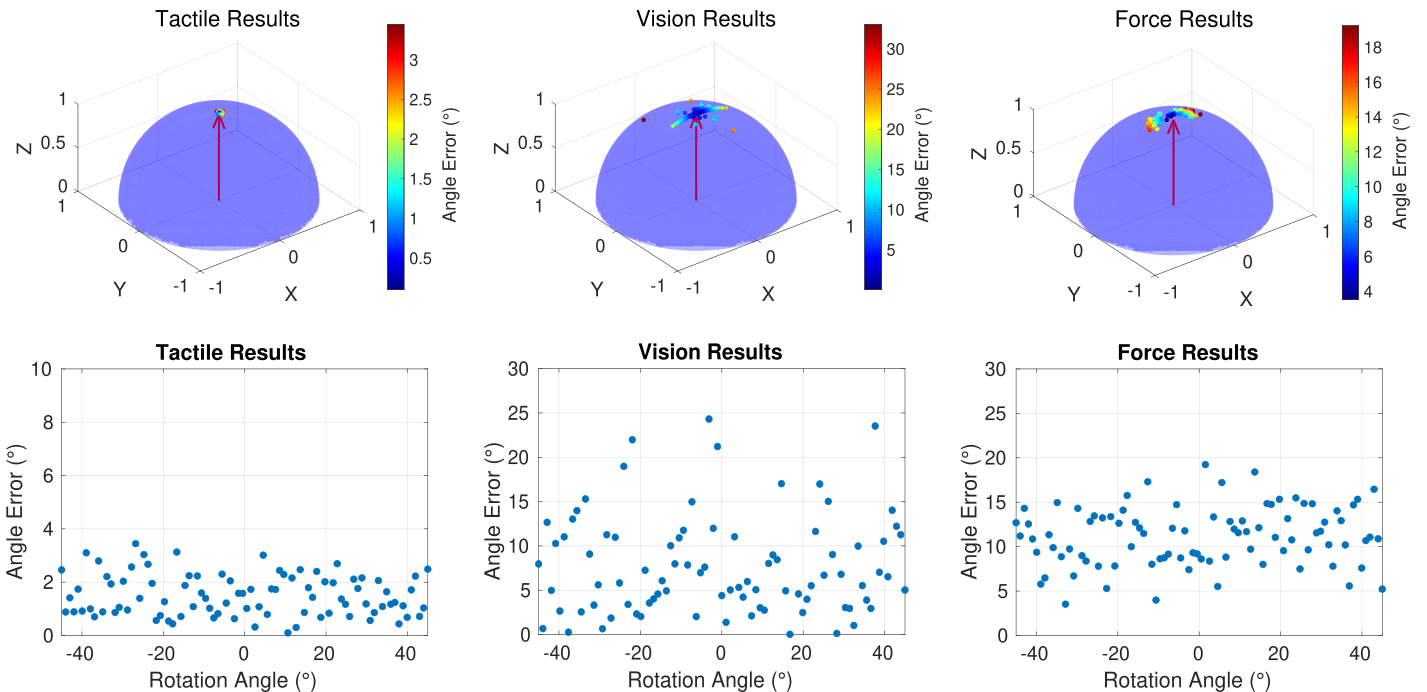}
    \caption{Experiments on contact plane estimation. \textbf{Top Row:} The visualisation of the normal vector of the estimated plane. \textbf{Bottom Row:} Quantitative plots show that contact planes estimated through tactile perception exhibit superior accuracy compared to those derived from visual and force information.}
    \label{fig: 9}
\end{figure*}

\subsubsection{Experiments on Plane Estimation} 
\begin{figure}
    \centering
    \includegraphics[width = \linewidth]{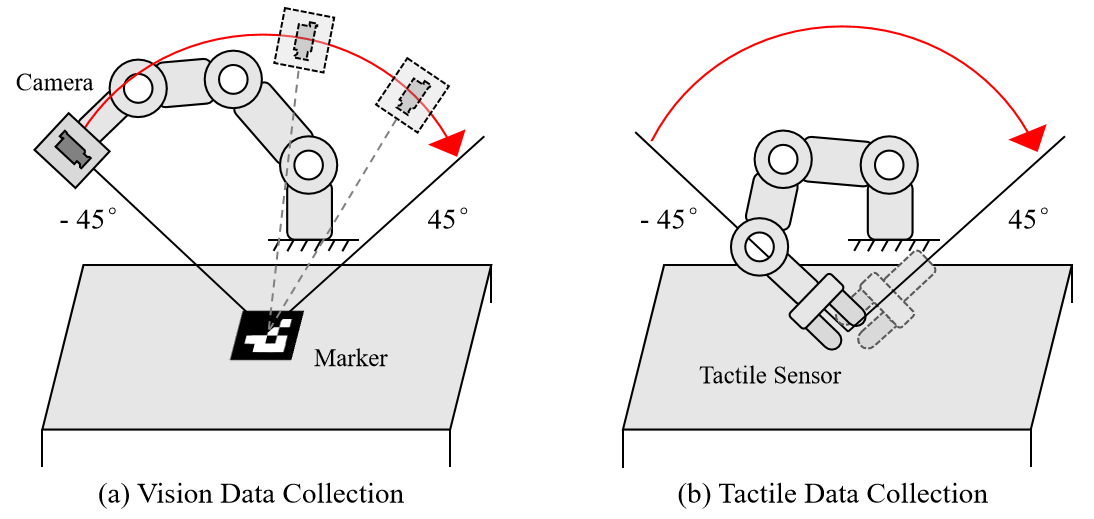}
    \caption{Sketch for \textbf{(a) visual data collection} and \textbf{(b) tactile and force data collection}. The rotation spans from -45$^\circ$ to 45$^\circ$, with each increment of 1$^\circ$.}
    \label{fig: 10}
\end{figure}

We evaluate the accuracy of three different sensing modalities: a vision-based system using ArUco markers in Sec.~\ref{Sec: grasp generation}, force sensing integrated into the UR5e robotic arm, and our proposed tactile contact plane estimation in Sec.~\ref{Section: Tactile-based Adjustment}. 
To facilitate this comparison, we collect the data from those different sensing modalities as shown in Fig.~\ref{fig: 10}. 

We adjust the rotation angle of the end-effector equipped with RoTip sensors, followed by remote observation of the ArUco Marker on the table and physical contact with it.
The rotation spans from -45$^\circ$ to 45$^\circ$, with each increment of 1$^\circ$. The ground truth for the normal contact plane remains parallel to the $z$-axis of the world frame, as the marker is taped to the table upon which the robotic arm's base is situated. Hence, the ground truth is represented by the vector $\left(0, 0, 1\right)$, as shown in the first row of Fig.~\ref{fig: 9}. The accuracy is evaluated through the angle error $\mathbf{e_n}$ between the ground truth normal vector $\mathbf{n}_{\text {real }}$ and the estimated normal vector $\mathbf{n}_{\text {estimated}}$ of the plane, computed as follows: 
\begin{equation}
    \mathbf{e_n} =\arccos \left(\mathbf{n}_{\text {estimated}} \cdot \mathbf{n}_{\text {real }}\right).
    \label{eq: angle error}
\end{equation}  

Similar to the coordinate transformation in Eq.~\ref{eq: 7}, the ArUco marker coordinate estimated with vision can be calculated through $\boldsymbol{\mathrm{R}_W^M} = \boldsymbol{\mathrm{R}_W^E} * \boldsymbol{\mathrm{R}_E^C} * \boldsymbol{\mathrm{R}_C^M}$, where $\boldsymbol{\mathrm{R}_A^B}$ represents the rotation matrix from $\boldsymbol{A}$ coordinate to $\boldsymbol{B}$ coordinate. The third column of $\boldsymbol{\mathrm{R}_W^M}$ represents the plane surface normal vector in the world frame.  
To estimate the plane normal with the UR5e integrated force sensor, we first obtain the wrench data from the UR5e's ROS (Robot Operating System) driver. Then, we transform the wrench data referenced in the end-effector coordinate to the world coordinate.

In the top row of Fig.~\ref{fig: 9}, each point represents a normalised surface vector estimated with corresponding sensing modalities. A more concentrated distribution of these points indicates a higher accuracy in the estimation results.  
Notably, our RoTip sensor demonstrates significantly higher accuracy in sensing the plane surface compared to both the force sensing and the vision system. 
We analysed the angle error using Eq.~\ref{eq: angle error}, revealing that our approach achieves accurate plane estimation with an average error of 1.51$^\circ$. It significantly outperforms vision system-based and integrated force sensing methods, which have an average error of 7.82$^\circ$ and 11.14$^\circ$, respectively. Additionally, the visualisation of angle errors in the bottom row of Fig.~\ref{fig: 9} confirms the superior accuracy of our tactile plane estimation approach.

Similar to the approach described in~\cite{flipbot}, we restrict our use of depth information to the area surrounding the corners of the object. Hence, the primary sources of error in the vision system are attributed to noise within the depth-fitted plane and insufficient depth information. However, the accuracy is limited by the marker’s size and the quality of depth data from the RealSense camera. This could be addressed by utilising depth information from the entire table surface or employing a higher-resolution depth sensor.

On the other hand, the discrepancy of the force sensing likely arises from the accuracy of the integrated F/T sensor. The RoTip sensors and the control board attached to the end-effector add weight to the robotic arm, which can introduce bias into the force readings. While the UR5e robotic arm includes a calibration kit to estimate the mass of attached loads and mitigate these effects, this calibration could be challenging and the gravitational load can therefore introduce variations in the readings from the force-torque sensor, which may affect the precision of plane fitting. Although using a more advanced force sensor might reduce these estimation errors, it would increase costs. Thus, our RoTip sensors offer an affordable solution for handling thin and flexible objects while maintaining reliable sensing performance.

\subsubsection{Experiments on Two-finger Contact}
Subsequently, we evaluate the success rate of achieving contact after adjusting the end-effector pose using various plane fitting approaches. A successful contact is defined by the contact areas detected by the RoTip sensor exceeding a predefined threshold. Since the Robotiq gripper is equipped with two RoTip sensors, our evaluation includes both the success rate of single-finger contact and two-finger contact.

We tested four different plane estimation approaches with 20 trials each to evaluate the contact success rate, as shown in Table~\ref{tab: contact experiment}. 
The ``Vision" method directly guides the end-effector to the target position. The ``Vision + Force (WA)" method positions the end-effector to a pre-contact pose and establishes normal contact, halting movement when the force exceeds a certain threshold. The ``Vision + Force" method refines the target pose by integrating plane estimation from force sensing. Our proposed ``Vision + Force +Tactile" method adjusts the target pose based on the results of tactile plane estimation and uses force measurements to control the robot.

\begin{table}
	\begin{center}
	
		\caption{\footnotesize The success rates of two-finger contact experiments. }
		\label{tab: contact experiment}
        \scalebox{1.0}{
		\begin{tabular}{c | c | c  }
			\hline
			Approach  & One-Finger SR & Two-Finger SR    \\
			\hline
			\hline
% 			{=|=|=|=|=|=|=}
		      Vision &  0.50 & 0.30  \\
            Vision + Force (WA) &  1.0 & 0.50 \\
    		Vision + Force &  1.0 & 0.55 \\
            Vision + Force + Tactile (Ours) &\textbf{1.0}  & \textbf{1.0} \\
            \hline
		\end{tabular}}
        % \newline
        % WA: Without Adjustment. \qquad \qquad \qquad \qquad \qquad \qquad \qquad \qquad
        \end{center}
        \quad WA: Without Adjustment.
        
\end{table}
As reported in Table~\ref{tab: contact experiment}, achieving two-finger contact relying solely on vision is challenging. Even one-finger contact cannot be ensured due to translation errors in grasp pose estimation. At times, the end-effector is positioned slightly away from the object, while at other times, it is too close, triggering an emergency stop. The ``Vision + Force (WA)" and ``Vision + Force" methods improve the reliability of one-finger contact but lack the precision needed for two-finger contact. Although the vision-based method has a lower average plane fitting error, the plane normal error in the $y$-axis significantly affects the height difference between the fingertips of two RoTip sensors, impacting the success rate of contact. As shown in the top row of Fig.~\ref{fig: 9}, the $y$-axis error for plane estimation using force data is similar to that of vision, resulting in comparable contact success rates. Our proposed method, which integrates vision, force sensing and tactile sensing, not only ensures one-finger contact but also significantly enhances the accuracy of achieving two-finger contact, with a success rate of 100\%. This two-finger contact is crucial for handling thin and flexible objects: If the passive finger used to hold the object loses contact, the object will slide on the table during the rotation of the active finger. In contrast, if the active finger used for rotation loses contact, the object cannot be manipulated.

\begin{figure}[t]
    \centering
    \includegraphics[width=1.0\linewidth]{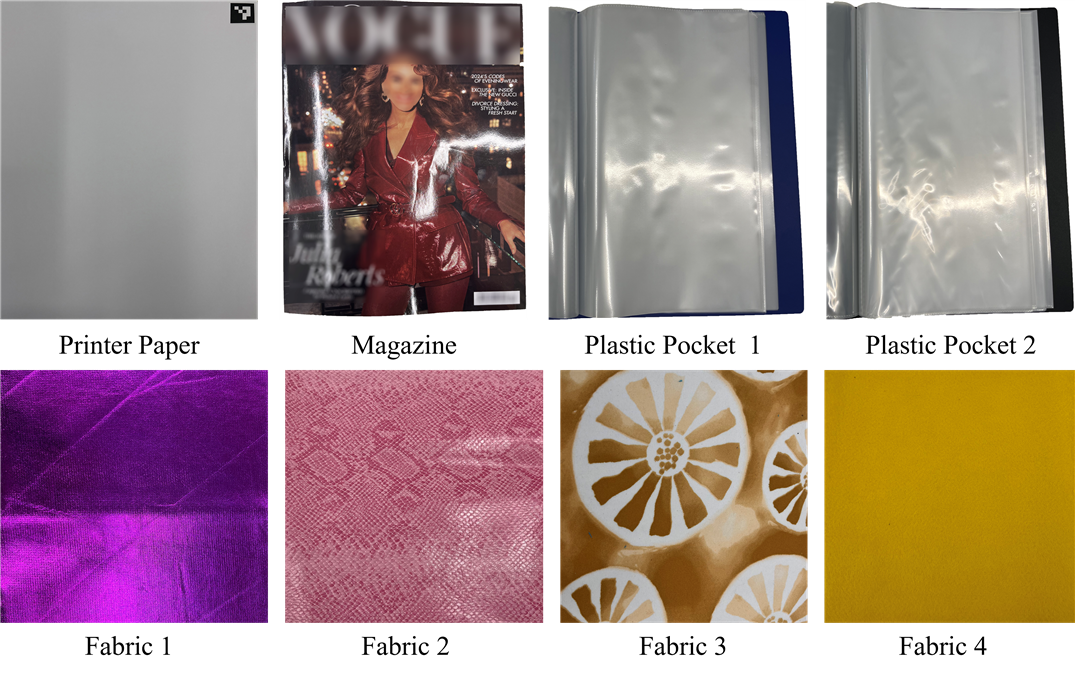}
    \caption{Objects used in our experiments. \textbf{From left to right:} Top row includes printer paper, coated paper (magazine) and two plastic pocket sheets for comparing against the SOTA methods~\cite{jiang2019dynamic, flipbot}. The bottom row includes four fabrics with different stiffness and textures, used to test the generalisation ability of our proposed approach.}
    \label{fig: 11}
\end{figure}

\subsection{Continuous Adjustment and Tactile Counting}
In this subsection, we evaluate the importance of the continuous adjustment and tactile counting proposed in Sec.~\ref{Section: Continuous Adjustment}. 
The objects used in this study are shown in Fig.~\ref{fig: 11}. Among these, printer paper, coated paper and plastic sheets are used for the comparison against the SOTA methods, while four different fabrics are used in the test of generalisation ability of our method. Printer paper exhibits the highest elasticity, coated paper (magazine) has the lowest elasticity, and plastic pocket sheets fall in between with medium elasticity.

\begin{figure}[t]
    \centering
    \includegraphics[width =\linewidth]{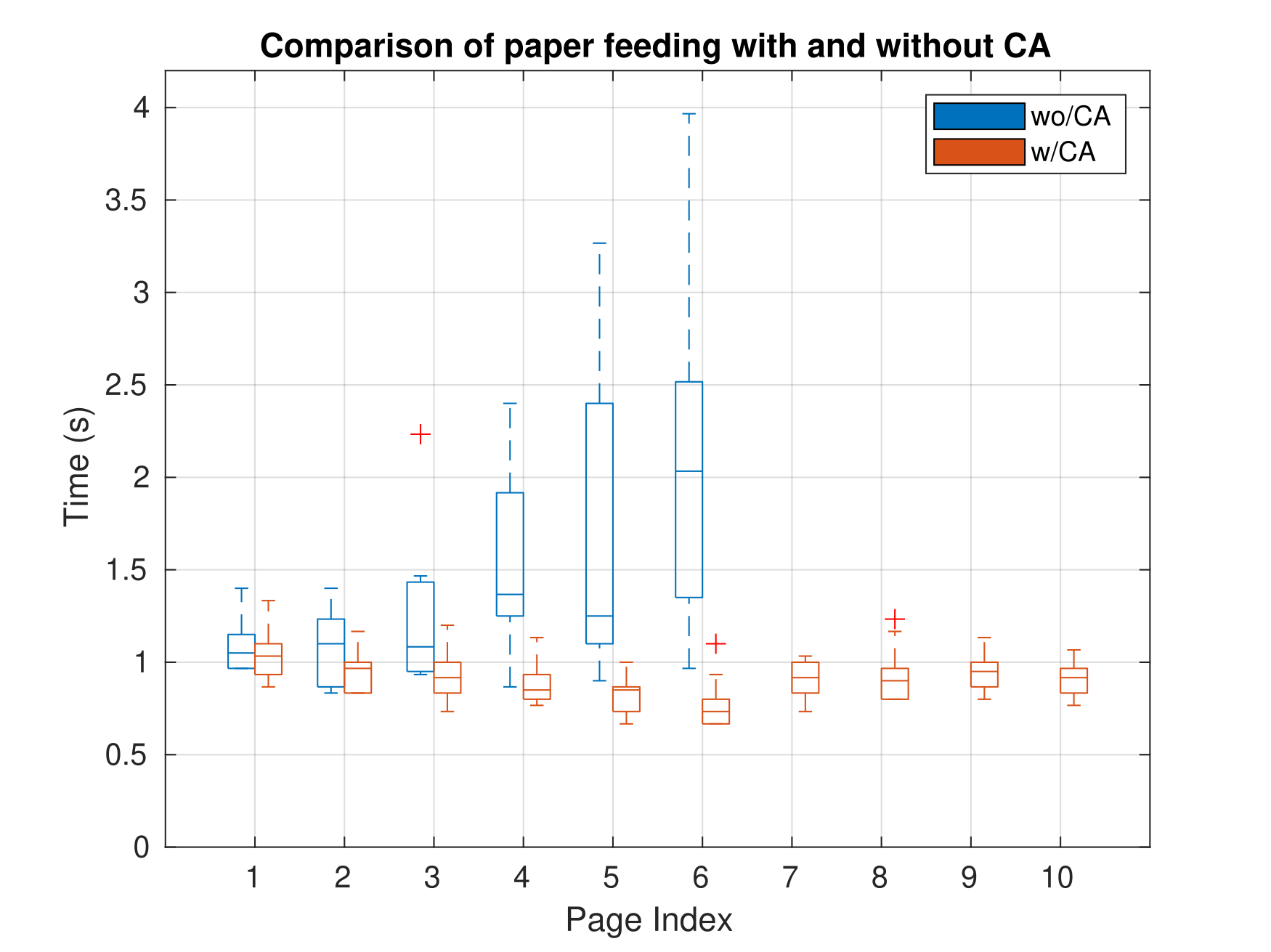}
    \caption{Box plot of the feeding time taken for each sheet, with and without continuous adjustment. Our proposed continuous adjustment maintains a consistently low feeding time for all sheets with minimal variance. In contrast, without continuous adjustment, the feeding time per sheet gradually increases with higher variance, and the system fails to feed beyond six sheets (hence, no plots for wo/CA beyond six sheets in the figure).}
    \label{fig: 12}
\end{figure}

\begin{table}[t]
    \begin{center}
        
    \caption{\footnotesize The time taken to feed each page.}
    \label{tab: times}
    \scalebox{1.0}{
    \begin{tabular}{c|c|c|c|c}
        \hline
        Page index &  1-3 &   4-6 &   7-9 &   10-12  \\
        \hline
        \hline
        wo/CA & 1.14\,\textpm\,0.30 & 1.78\,\textpm\,0.80 & N/A & N/A \\
        \hline
        w/CA &  0.98\,\textpm\,0.15 & 0.83\,\textpm\,0.13 & 0.93\,\textpm\,0.12 & 0.84\,\textpm\,0.13 \\
        \hline

    \end{tabular}}
    \end{center}
    % \centering
        
    \quad \quad CA: Continuous Adjustment.
        
\end{table}

\subsubsection{Experiments on Continuous Adjustment}
We first assess the importance of continuous adjustment. Specifically, we placed 15 sheets of printer paper on the table for feeding and grasping, and set the initial applied normal force of each finger to 4N. To minimise uncertainty in the visual system, objects were placed in predefined locations. The evaluation metric used here is the time required to feed each sheet.

We recorded the time required to feed each sheet with and without continuous adjustments. As shown in Table~\ref{tab: times}, for the first three sheets, when contact is sufficient, continuous adjustments have minimal effect. However, as more sheets are fed, contact becomes increasingly inadequate. Without continuous adjustment, the time to feed each sheet increases significantly, and only the first six sheets could be consistently fed. The robot fails to feed beyond six sheets (marked as N/A in Table~\ref{tab: times} for the 7-9 and 10-12 cases). In contrast, all sheets can be successfully fed when continuous adjustment is used. We also visualise the feeding time for each sheet with a box plot in Fig.~\ref{fig: 12}. 
This plot shows that our proposed RoTipBot with continuous adjustment maintains a consistent average feeding time and a small variance across all sheets, unlike the increasing feeding time and higher variance observed without continuous adjustment. This consistency is crucial for tasks that require precise processing and handling of multiple thin and flexible objects.

\begin{figure}
    \centering
     \includegraphics[width=1\linewidth]{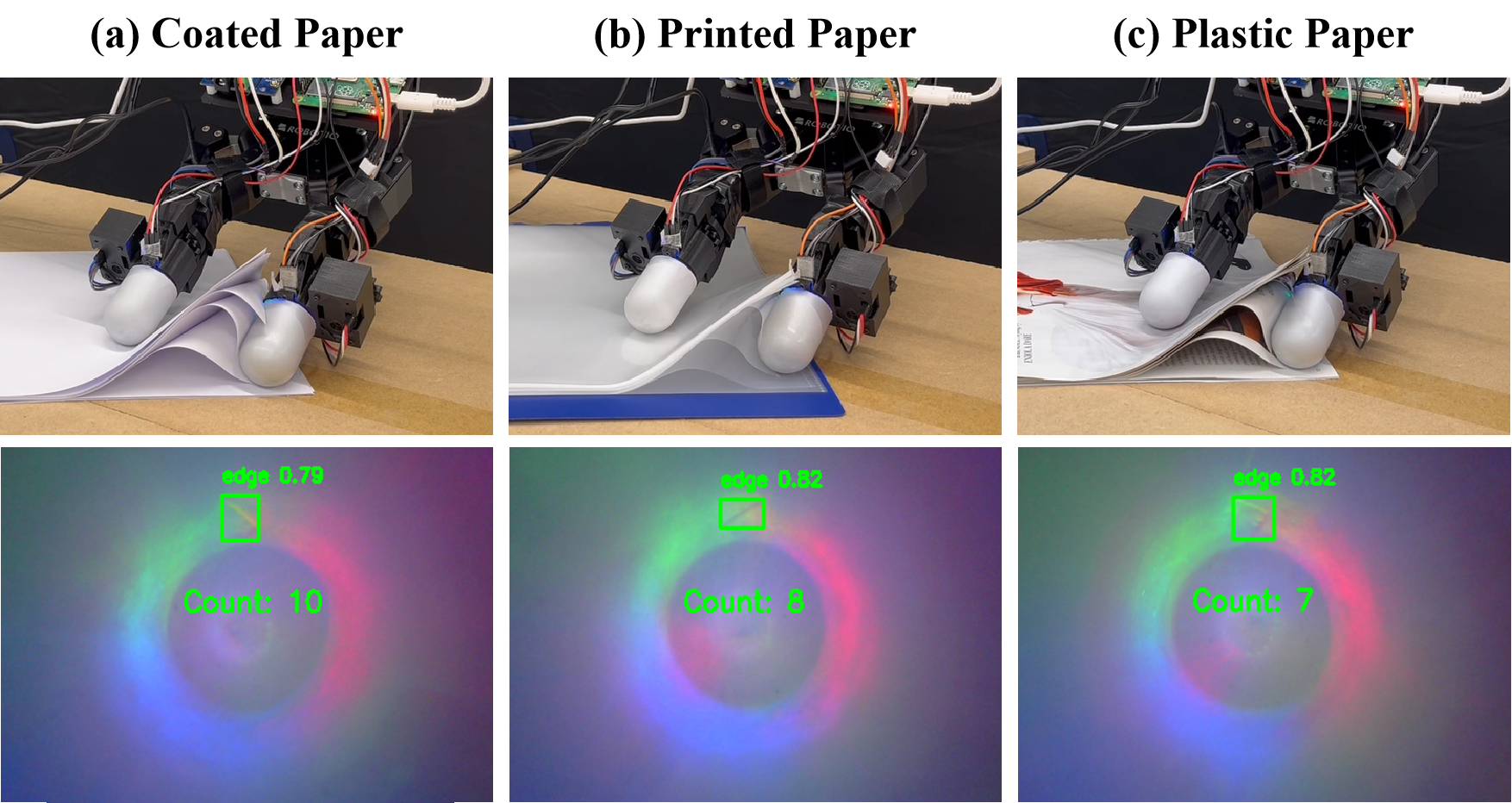}
    \caption{Snapshots of tactile counting experiments using three materials (printer paper, plastic sheet, and coated paper). }
    \label{fig: 13}
\end{figure}

\begin{table*}[t]
	\begin{center}
	\caption{\footnotesize Comparison of grasping Success Rate (SR) and the operation speed, i.e., Pages Per Minute (PPM).}
		\label{tab: comparison SOTA}
        \scalebox{1}{
		\begin{tabular}{c| c| c | c | c | c| c| c| c | c| c}

		\hline   \multirow{2}{*}{Approach} &  \multirow{2}{*}{Angle} & \multirow{2}{*}{Multi-page Grasping} & \multicolumn{2}{c|}{Printer paper} & \multicolumn{2}{c|}{Coated paper} & \multicolumn{2}{c|}{Plastic pocket sheet} &  \multicolumn{2}{c}{All}\\
                \cline{4-11}
			  &   &  & SR $\uparrow$ &  PPM $\uparrow$  &  SR $\uparrow$ &  PPM $\uparrow$  &  SR $\uparrow$ &  PPM $\uparrow$ & SR $\uparrow$ & PPM $\uparrow$ \\
			\hline
% 			{=|=|=|=|=|=|=}
    		\hline
                Flex\&Flip~\cite{jiang2019dynamic} & \multirow{4}{*}{60$^\circ$} & $\times$ & 60\% & 3.10 &57\% &2.95 & 43\% &2.59 & 53.3\% & 2.88 \\
                FlipBot~\cite{flipbot} &  & $\times$ & 83\% & 4.28 & 81\% & 4.18 & 68\% & 3.52 & 77.3\% & 3.99 \\
                % \cline{3-11}
                RoTipBot (w/o counting) &  & $\checkmark$  & 90\% & \textbf{12.3} & 87\% & \textbf{10.9} & 94\% & \textbf{11.3} & 90.3\% & \textbf{11.5} \\
                RoTipBot (Ours) &  & $\checkmark$  & \textbf{95}\% & 11.0 & \textbf{91}\% & 9.9 & \textbf{97}\% & 10.9 & \textbf{94.3}\% & 10.6 \\
                
                \hline
                Flex\&Flip~\cite{jiang2019dynamic} & \multirow{4}{*}{30$^\circ$} &  $\times$ & 69\% & 3.57 & 73\% & 3.77 & 43\%  & 2.22  & 61.7\% & 3.19\\
                FlipBot~\cite{flipbot} & & $\times$ & 91\% & 4.68 & 91\% & 4.70 & 67\% & 3.46 & 83.0\% & 4.28\\
                % \cline{3-11}
                RoTipBot (w/o counting)  & &  $\checkmark$ & 92\% & \textbf{11.7} & 89\% & \textbf{11.2} & 93\% & \textbf{11.5} & 91.3\% & \textbf{11.5} \\
                RoTipBot (Ours) &  & $\checkmark$  & \textbf{95}\% &10.5 & \textbf{92}\% & 10.7 & \textbf{98}\% & 10.4 & \textbf{95.0}\% & 10.5 \\
                                
                \hline
                Flex\&Flip~\cite{jiang2019dynamic} & \multirow{4}{*}{0$^\circ$} & $\times$ & 71\% & 3.65& 80\% & 4.11 & 51\% & 2.61 & 67.3\% & 3.46 \\
                FlipBot~\cite{flipbot} & & $\times$ & 92\% & 4.75 & \textbf{95}\% & 4.91 & 75\% & 3.88 & 87.3\% & 4.51\\
                % \cline{3-11}
                RoTipBot (w/o counting) &  &  $\checkmark$ & 92\% & \textbf{12.1} & 90\% & \textbf{11.5} & 94\% & \textbf{11.8} & 92.0\% & \textbf{11.8}\\
                RoTipBot (Ours) &  & $\checkmark$  & \textbf{96}\% & 11.2 & 93\% & 10.2 & \textbf{97}\% & 10.7 & \textbf{95.3}\% & 10.7 \\

                \hline

		\end{tabular}}
        \end{center} 
        \quad \quad \quad \quad Note: The results for FlipBot~\cite{flipbot} and Flex\&Flip~\cite{jiang2019dynamic} are copied from the SOTA FlipBot paper~\cite{flipbot}.

\end{table*}

\begin{table}[t]
    \begin{center}
        
    \caption{ The accuracy of counting and bounding box detection for paper boundaries.}
    \label{tab: counting}
    \scalebox{1.0}{
    \begin{tabular}{c|c|c|c}
        \hline
        Object &  Print Paper &   Coated Paper &  Plastic Sheet   \\
        \hline
        \hline
        Recall &0.919  & 0.878& 0.954\\
        \hline
        AP50 &  0.945 & 0.941 & 0.981\\
        \hline
        AP50-95 & 0.553 & 0.497& 0.573\\
        \hline 
        Counting Accuracy & 97\% & 94\% & 98\%\\
        \hline
    \end{tabular}}
    \end{center}
    % \centering
        
\end{table}

\subsubsection{Experiments on Tactile-based Object Counting}
We evaluate the object counting performance from two key perspectives: edge detection accuracy and counting accuracy. The results are summarised in Table~\ref{tab: counting}, which presents the accuracy of counting and bounding box detection for paper boundaries across different materials. The table shows metrics such as Recall, AP50 (Average Precision at 50\% Intersection over Union threshold), AP50-95 (Average Precision across IoU thresholds from 50\% to 95\%), and counting accuracy for three types of objects, i.e., print paper, coated paper, and plastic sheet. 

The results summarised in Table~\ref{tab: counting} demonstrate strong performance in boundary detection and counting across all materials, with plastic sheets achieving the highest Recall (0.954) and AP50 (0.981), followed by print papers and then coated papers. AP50-95 scores, which evaluate precision across a range of IoU thresholds, indicate consistent accuracy but show a slight decrease for coated papers (0.497), likely due to its softness and smallest thickness, which can affect detection precision.

Counting accuracy remains high for all materials, underscoring the system’s reliability in diverse conditions. Specifically, print paper achieves 97\% accuracy, coated paper 94\%, and plastic sheet 98\%. The snapshots of tactile counting experiments using three materials are shown in Fig.~\ref{fig: 13}. 
Tactile-based object counting could be highly beneficial for tasks such as flipping to a specific page range in a book by detecting the number of pages through touch. This method relies on physical feedback, making it less dependent on lighting conditions, which provides a significant advantage in environments where book indices are obscured or unavailable.

\begin{figure*}
    \centering
    \includegraphics[width=\linewidth]{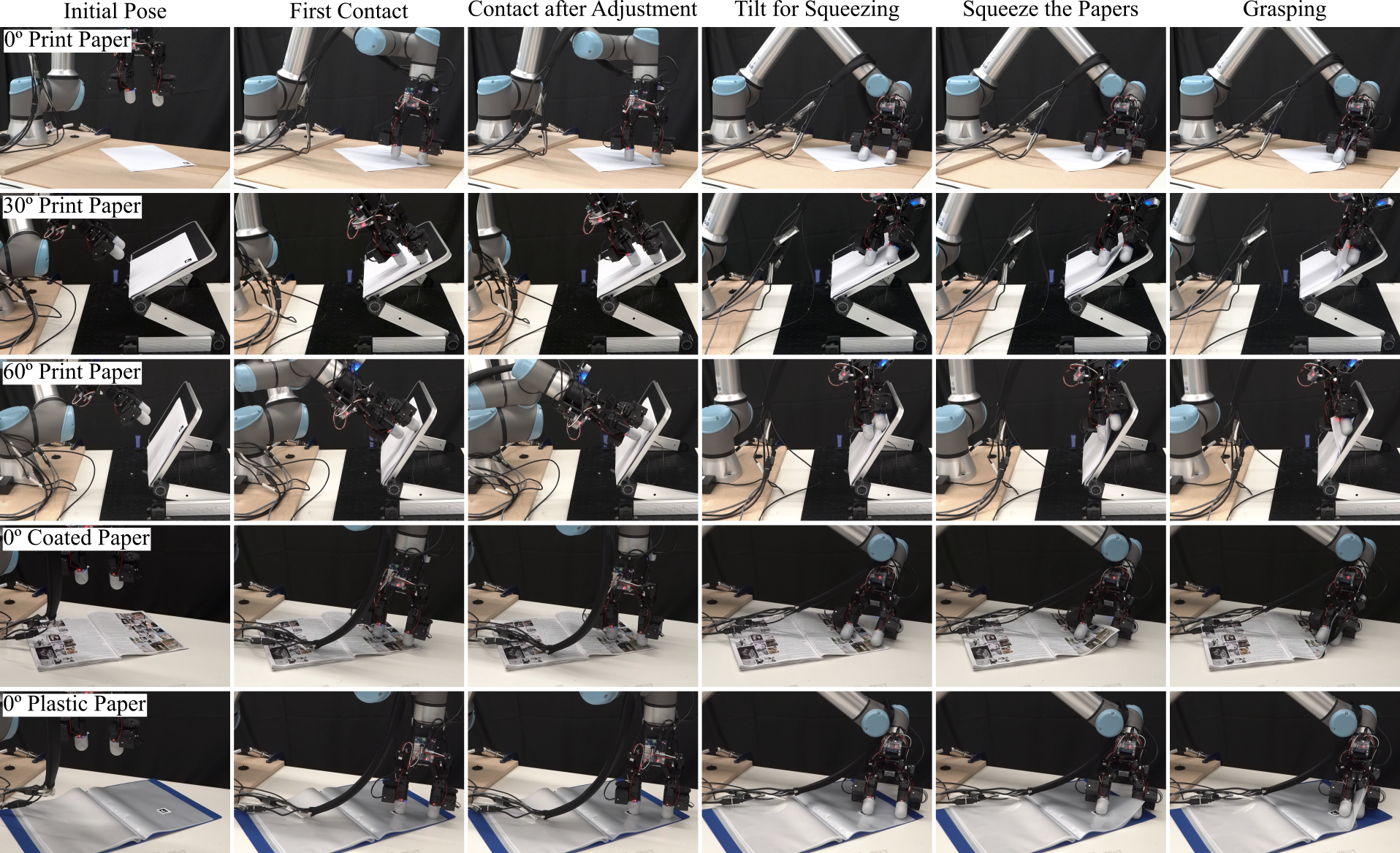}

    \caption{Snapshots of the contact and grasping stages. The first three rows show the grasping of printer papers at 0$^\circ$, 30$^\circ$ and 60$^\circ$. The last two rows show the grasping of coated papers and plastic pocket sheets. For each row, from left to right: initial pose, first one-finger contact, two-finger contact after tactile-based adjustment, tilting to feed the paper, feeding the papers, and closing the gripper to grasp. }
    \label{fig: 14}
\end{figure*}

\subsection{Experiments on Thin and Flexible Object Handling}
In this subsection, we conduct comprehensive experiments to evaluate our proposed RoTipBot approach for handling thin and flexible objects. 
First, we compare our proposed method against the state-of-the-art (SOTA) methods~\cite{jiang2019dynamic, flipbot}. Then, we conduct an ablation study to compare handling performance using different types of fingertips.
Finally, we test the generalisation ability of our method and showcase the failure cases during thin and flexible object handling.

\subsubsection{Comparison against SOTA}
Following the benchmarks set by the SOTA FlipBot approach~\cite{flipbot} and other multi-object grasping approach~\cite{agboh2022multi}, we evaluate our proposed approach using both Success Rate (SR) and Pages Per Minute (PPM) metrics, providing a comprehensive assessment of its effectiveness and efficiency. Unlike FlipBot~\cite{flipbot}, which is limited to grasping a single sheet at a time, our RoTipBot can grasp multiple sheets in one trial. Hence, we define the SR as follows:
\begin{equation}
   \text{SR} = 1 - \frac{\left| \text{Grasped Number} - \text{Target Number} \right|}{\text{Target Number}}
\end{equation}

In this experiment, ten repetitions for each trial with a target set at ten pages are conducted. Three types of thin and flexible objects with different physical properties were used, i.e., printer paper, coated paper, and plastic pocket sheet, consistent with those used in~\cite{flipbot}. This selection allowed for a direct comparison of our RoTipBot against SOTA methods. To test the effect of gravity on paper flipping, we placed the object on a laptop stand, as shown in Fig.~\ref{fig: 3}, and varied its tilt angles at 0, 30, and 60 degrees.

As shown in Table~\ref{tab: comparison SOTA}, our RoTipBot consistently achieves more than a 90\% success rate regardless of the tilt angle. Moreover, RoTipBot outperforms the SOTA method~\cite{flipbot} in terms of overall success rate across various angles, achieving up to a 17\% improvement at 60$^\circ$ and at least a 8\% improvement at 0$^\circ$. 
More importantly, RoTipBot operates significantly faster than the SOTA approaches, achieving speeds up to three times faster. We also show snapshots of grasping different objects at various tilt angles in Fig.~\ref{fig: 14} for visualisation.

We also compare the results of our RoTipBot approach with and without the tactile counting feature. In the baseline approach, the stopping condition for each finger rotation is set by a fixed time per layer $t_{feed}=l_c/v_{feed}$, where $l_c$ is the distance between the RoTip sensor and the object corner as shown in Fig.~\ref{fig: 4}, and $v_{feed}$ is the linear velocity at the contact position between RoTipBot and the paper during the feeding process. The total estimated grasping time for  $n_{desired}$ layers is then $n_{desired}*t_{feed}$.
We find that the incorporation of tactile counting improves grasping accuracy, although it slightly reduces efficiency. This trade-off arises because faster rotations lead to more blurred tactile images, which reduces the time available for accurate counting. To optimise the performance, we adjusted the finger rotation speed to be slightly slower, striking a balance that improves grasping accuracy while still maintaining a significantly faster operation than the baseline method.

Specifically for different objects, our method modestly outperforms the SOTA with printed papers and significantly enhances success rates for plastic pocket sheets, showing a minimum improvement of 22\%, across different angles. However, coated papers present the most significant challenge, with slightly lower performance compared to the SOTA when grasping at 0$^\circ$.
This performance gap deviates from the results of FlipBot~\cite{flipbot}, where our method excels at grasping plastic pocket sheets compared to coated paper, contrary to the trends observed in FlipBot. The challenge with coated papers stems from their small thickness and lack of elasticity, making them hard to be detected by the RoTip sensing during counting and more susceptible to gravity. This either causes coated papers to be miscounted, or makes it difficult to feed them completely between the two RoTip sensors. In contrast, printer papers and plastic pocket sheets have stronger elasticity and larger thickness, which helps avoid creasing issues. However, ArUco markers printed on printer papers can introduce static electricity, complicating separation and reducing grasping performance compared to plastic sheets. Moreover, tiny protrusions on the edges of plastic sheets create small gaps between layers, preventing van der Waals forces from hindering movement~\cite{dzyaloshinskii1961general}. This makes separation easier and facilitates more efficient grasping.

\subsubsection{Ablation Study}
We conducted an ablation study to examine the impact of tactile feedback, continuous adjustment, and finger stiffness on manipulation performance, while preserving the finger’s rotational freedom. We evaluated four configurations, each differing in tactile adjustment, continuous adjustment, and finger stiffness:
\begin{itemize}
    \item Case 1 (RoTipBot): with continuous adjustment,  tactile adjustment and soft elastomer skin.
    \item Case 2 (TAS): with Tactile Adjustment and Soft elastomer skin only.
    \item Case 3 (S): with Soft elastomer skin only.
    \item Case 4 (-): without soft elastomer skin, though it includes the same coating to preserve surface properties, and without tactile adjustment and continuous adjustment.
\end{itemize}

As summarised in Table~\ref{table:ablation_results}, Case 1 (RoTipBot) achieved the highest success rate of 94.9\%, demonstrating that the combination of soft elastomer skins with both tactile adjustment and continuous adjustment significantly improves manipulation performance. Removing continuous adjustment in Case 2 reduced the success rate to 63.7\%. While tactile sensing for sufficient contact still contributed to grasping accuracy, the lack of continuous adjustment hindered the system's adaptability to changes in paper thickness. In Case 3, the exclusion of both tactile and continuous adjustments further decreased the success rate to 28.2\%, as vision alone proved insufficient for precise paper handling. Finally, the rigid-finger configuration (Case 4) failed completely, as the lack of soft elastomer limited the contact area, significantly reducing friction and preventing effective interaction with the paper. This ablation study highlights the critical role of RoTipBot's integrated components—tactile adjustment, continuous adjustment, and soft elastomer skin—in achieving precise and reliable manipulation of thin, flexible objects.

\begin{table}[t]
\centering
\caption{Ablation study results showing the impact of soft skin, tactile adjustment (TA), and continuous adjustment (CA) on success rates.}
\begin{tabular}{c|c|c|c|c}
\hline
\textbf{Feature}           & \textbf{RoTipBot} & \textbf{CAS} & \textbf{S} & \textbf{-} \\ \hline
\hline
Soft skin                & \checkmark                     &\checkmark                &\checkmark                & $\times$                \\ \hline
TA         &\checkmark                       &\checkmark                & $\times$               & $\times$                \\ \hline
CA     &\checkmark                       & $\times$                & $\times$                & $\times$               \\ \hline
\textbf{Success Rate}      & \textbf{94.9\%}       & 63.7\%          & 28.2\%          & 0\%             \\ \hline
\end{tabular}

\label{table:ablation_results}
\end{table}

% In this ablation study, the soft fingertip with tactile sensing demonstrates the highest success rate and handling stability, particularly in xx. The soft fingertip without tactile sensing performs moderately well but lacks responsiveness to xx. The rigid fingertip without tactile sensing shows the lowest performance, with limited xx.
% adaptability and stability during handling tasks.

\begin{figure*}[t]
    \centering
    \includegraphics[width=\linewidth]{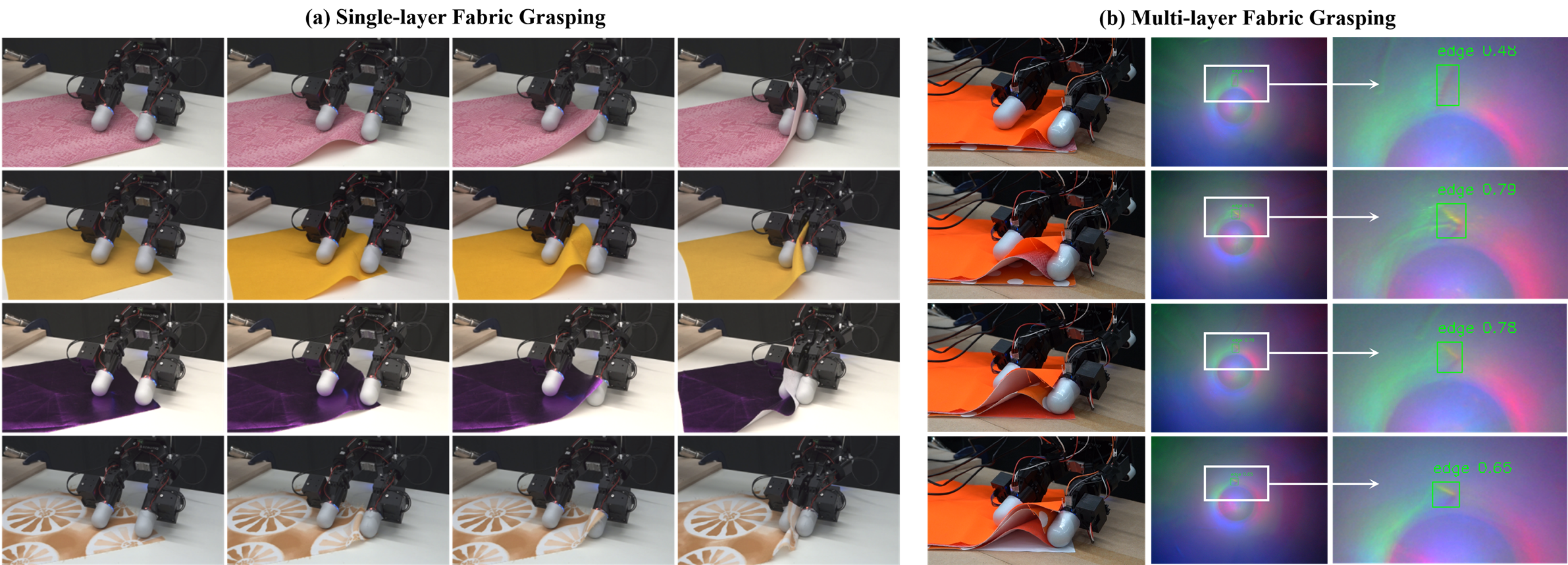}
    \caption{(a) Snapshots of grasping four different single-layer fabrics. \textbf{First column:} The Robotiq gripper equipped with RoTip sensors first moves down to touch the fabric. \textbf{Second and third columns:} Then, one RoTip sensor squeezes the fabric to feed it between the two RoTip sensors. \textbf{Last column:} Finally, the gripper is closed to hold the fabric. (b) Snapshots of grasping multi-layer fabrics. \textbf{First column:} Grasping of different layers. \textbf{Second column:} Tactile feedback while grasping. \textbf{Third column:} Cropped tactile images of the contact area.}
    \label{fig: 15}
\end{figure*}

\subsubsection{Generalisation and Failure Cases}
The generalisation of our RoTipBot was further evaluated through tests on four distinct types of fabrics as shown in the bottom row of Fig.~\ref{fig: 11}. RoTipBot achieved a nearly perfect success rate in grasping across various fabric types, approaching 100\%, as demonstrated in Fig.~\ref{fig: 15}-(a). We also demonstrate the ability to pick up multiple fabrics as shown in Fig.~\ref{fig: 15}-(b).

\begin{figure}
    \centering
    \includegraphics[width=\linewidth]{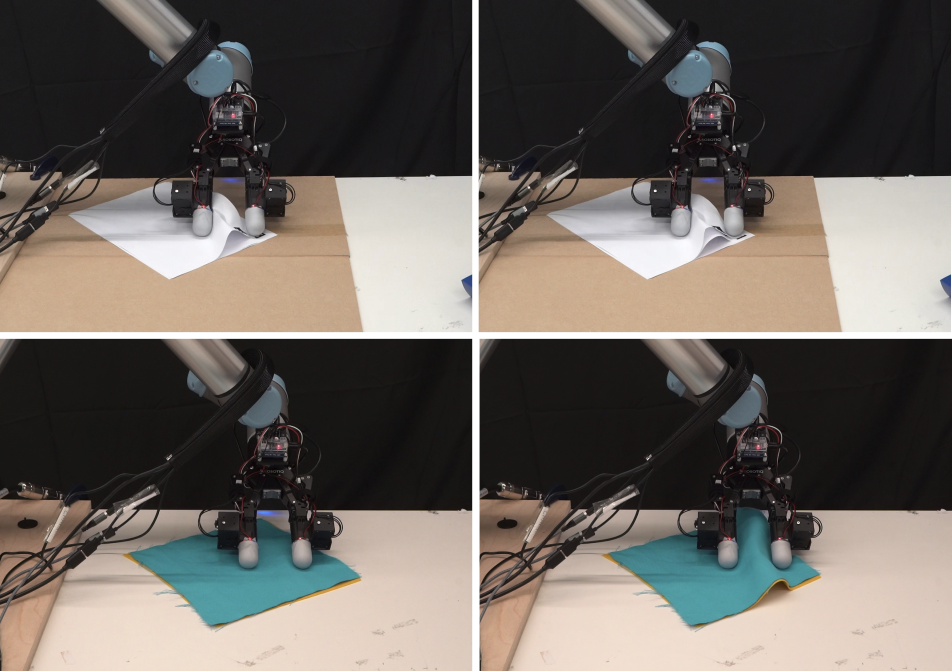}
    \caption{\textbf{Visualisation of Failures Cases.} The first row shows that the printer paper with wrinkles is hard to squeeze and stops the next page's movement, eventually leading to a feeding failure. The second row demonstrates a failure case when grasping objects with a high friction coefficient, where two layers of fabrics are squeezed together.}
    \label{fig: 16}
\end{figure}

However, handling multiple layers of fabric remains challenging, especially when there is significant variation in thickness and high friction between layers. The varying thickness can cause issues with continuous adjustments, as we assume a large thickness in multi-layer fabric grasping to maintain consistent contact between the finger and fabric. Additionally, high friction can lead to compression and bending of multiple layers, causing them to stick together and complicating the separation process. To achieve stable grasping and manipulation of such fabrics, a reinforcement learning approach could be used for maintaining continuous contact between the finger and the object. While our experiments demonstrated the effectiveness of our RoTipBot across a spectrum of applications, our real-world trials also brought to light some failure cases, as outlined below:

    \textbf{Grasping objects with a high friction coefficient.} When the friction coefficient between the RoTip sensor and the object's top surface is lower than that between the object's layers, the finger's squeezing action may not generate sufficient force to separate the top layer from the remaining layers. Furthermore, if the friction coefficient between the object and the table is lower than that between the finger and the object, multiple layers of the object may be squeezed and bent together, causing them to remain stuck to each other. This issue is particularly evident when handling rough fabrics consisting of multiple layers, as illustrated in the bottom row of Fig.~\ref{fig: 16}. A rotatable tactile sensor with high friction or suction capability could help overcome the difficulties posed by high friction coefficients, ensuring effective handling even when friction between layers is substantial.

    \textbf{Inadequate force applied to objects with wrinkles.} When handling stiff objects with wrinkles, such as crumpled printer papers and coated papers, the force distribution changes, leading to inadequate force being applied during manipulation. This issue is similar to a cash counting machine failing to process wrinkled currency. As shown in Fig.~\ref{fig: 16} (top row), wrinkled printer paper is difficult to squeeze, impeding the movement of the subsequent pages and causing handling failures. Integrating the RoTip sensors into a dexterous hand with a third finger could resolve this by applying additional force to manage the stuck paper.

\section{Discussion} \label{Section:6}

\subsection{Discussion on Rotatable Tactile Sensors}
In recent years, advancements in vision-based tactile sensing have primarily centred on enhancing sensing precision and expanding the range of sensing modalities. There has been a movement towards making tactile sensors more dynamic or mobilised to meet the requirements for more skilful manipulation. This includes dynamically adjusting the camera's position~\cite{lin2023gelfinger} or moving the sensor's surface via rotation and translation motions~\cite{cao2023touchroller,shimonomura2021detection}. Specifically, tactile sensors that can rotate utilise two distinct strategies: passive rotation and active rotation, as summarised in Table~\ref{tab: tactile sensors}.

Passive rotatable tactile sensors~\cite{shimonomura2021detection,cao2023touchroller} rely on the sensor's ability to adjust its orientation in response to external forces without an active mechanism. This method is characterised by simplicity, low cost, and high efficiency, and is particularly suitable for environments where these attributes are prioritised, such as exploration and monitoring in harsh environments.

Active rotatable tactile sensors~\cite{tacrot, lin2023gelfinger,yuan2023tactile,xu2024dtactive} involve the sensor itself moving, usually driven by an external actuator or motor. This allows for precise control of contact status and delicate manipulation of objects.
This active movement is particularly advantageous in robotic applications requiring detailed interaction, such as assembly lines or household care. However, it introduces challenges like increased complexity, higher costs, and greater power consumption. 

Compared to the previous active rotatable tactile sensors such as TacRot~\cite{tacrot} and TRRG~\cite{yuan2023tactile} that can only sense the inner side of the fingertip, our RoTip sensor is the first to provide tactile sensing across the entire fingertip area while enabling finger rotation. This mobilised omnidirectional tactile sensing significantly enhances dexterity. Consequently, RoTipBot is the first system to facilitate the grasping of thin and flexible objects with mobilised vision-based tactile sensing.

\begin{table}
    \centering
    \caption{\footnotesize Comparison of passive and active rotatable tactile sensors.}
    \label{tab: tactile sensors}
    \begin{tabular}{>{\centering}m{0.12\linewidth} | >{\centering}m{0.35\linewidth} | >{\centering\arraybackslash}m{0.35\linewidth}}
       \hline
       \textbf{} & \textbf{Passive Rotatable} & \textbf{Active Rotatable} \\
       \hline\hline
       \textbf{Pros} & Cost-effective, simple to implement & Precise control, applicable in a wide range of applications \\
       \hline
       \textbf{Cons} & Requires additional devices for full functionality & High complexity, increased costs \\
       \hline
       \textbf{Examples} & TouchRoller~\cite{cao2023touchroller}, Roller Tactile Sensor~\cite{shimonomura2021detection} & TacRot~\cite{tacrot}, TRRG~\cite{yuan2023tactile}, GelFinger~\cite{lin2023gelfinger}, RoTip (ours)
       \\
       \hline
    \end{tabular}
\end{table}

\subsection{Handling Policy for Thin and Flexible Objects} 
We employed the FEM due to its accuracy and reliability to determine the best locations for handling thin and flexible objects. FEM excels in modelling and simulating the physical response of materials to various forces. Nonetheless, it also has certain limitations. However, it has certain limitations, particularly the significant computational resources required, especially when dealing with complex materials such as thin and flexible objects.  

Conversely, Reinforcement Learning (RL) has gained attention as an effective method for interacting with objects and learning optimal behaviours through trial and error~\cite{flipbot}. Unlike FEA, RL does not need exhaustive pre-defined material properties and can improve its accuracy with increased exposure to diverse situations. This adaptability allows RL to adjust to new objects and scenarios over time. However, RL usually starts with lower accuracy and requires significant data and computational investment during the learning phase.

In the future, it would be interesting to develop a hybrid approach that combines the accuracy of FEM with the adaptability and learning capabilities of RL. Such an integration could leverage FEM for initial detailed simulations, providing a foundational understanding that can be further refined and adapted through RL's interaction-based learning. This hybrid model could address the limitations of both methods, leading to more robust and efficient systems for handling thin and flexible objects.

\subsection{Discussion on RoTip Sensors for Other Applications.}
Our study demonstrates the effectiveness of RoTip sensors in manipulating thin, flexible objects, while also highlighting their potential for broader applications in deformable object tasks, such as cable manipulation~\cite{she2021cable}. The active rotation capability of the RoTip sensors offers significant advantages over traditional sliding-based approaches. Unlike sliding, which can introduce unwanted twists or snags, active rotation enables precise control over a cable's orientation and tension, allowing smooth adjustments that minimise damage risk and enhance handling accuracy. This capability proves especially valuable in complex tasks such as routing, or dynamically positioning cables.

Moreover, the tactile sensing feature of RoTip provides an advanced level of perception for cable handling that is challenging to achieve with conventional vision-based methods. By continuously monitoring the contact surface, the sensor can detect subtle shifts in cable position, tension, and orientation in real time. This sensory feedback allows RoTip to adapt its grip and movements dynamically, a crucial feature for delicate operations such as managing wiring, fibre optics, or medical tubing. Together, the active rotation and tactile perception capabilities of RoTip enhance manipulation precision, enabling new possibilities for handling delicate, flexible materials in tasks that require both high precision and adaptability.

\begin{figure}
    \centering
    \includegraphics[width=\linewidth]{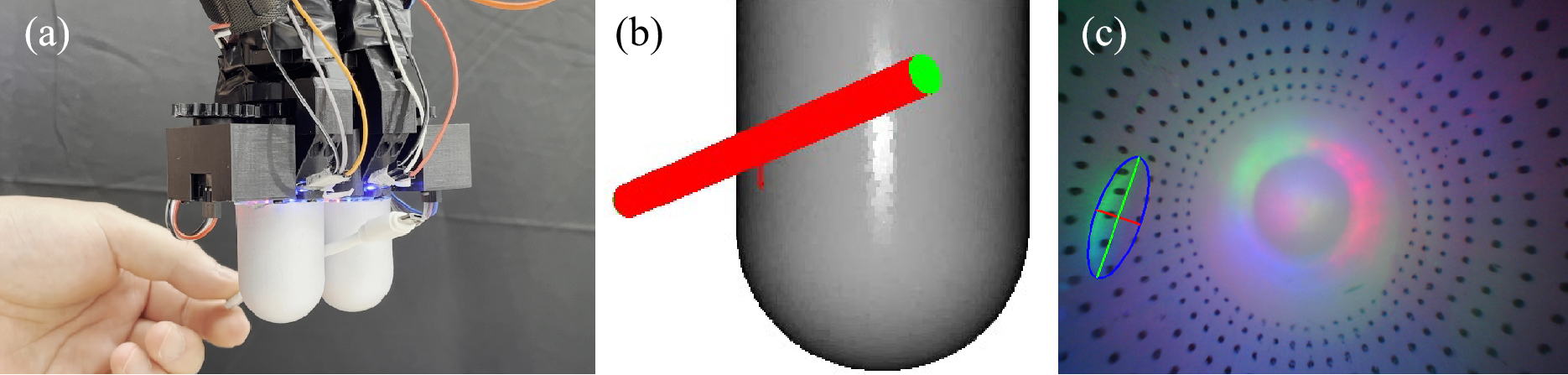}
    \caption{(a) Cable manipulation using the proposed RoTip sensor. (b) The estimated contact pose. (c) The corresponding tactile feedback.} 
    \label{fig: contact_pose}
\end{figure}

To demonstrate the sensor’s effectiveness in such applications, Fig.~\ref{fig: contact_pose} shows the relative pose estimation between the manipulated cable and the RoTip sensor. This capability provides real-time feedback on cable orientation, enabling the configuration to be centred and aligned with the fingers.

\subsection{Discussion on Tactile-based Adjustment}
Tactile-based adjustment, which involves adjusting the position and orientation of an object or a robot end-effector based on tactile feedback, has recently been used in various robotic applications~\cite{calandra2018more,  liang2023visuo}. For example, Hogan et al.~\cite{hogan2018tactile} designed a tactile-based policy for grasp adjustment, which adjusts the end-effector pose and improves the grasping quality. Jiang et al.~\cite{2022jiaqi} leveraged tactile feedback from a vision-predicted poking point predicted to refine the grasping pose of transparent objects. Lloyd and Lepora~\cite{ lloyd2023pose} used tactile feedback to adjust the end-effector's pose while exploring the object's surface. 

Many of these studies employ vision-based tactile sensors, recognised as cost-effective alternatives to expensive force sensors. For instance, the GelSight Mini\footnote{https://www.gelsight.com/gelsightmini/} and DIGIT~\cite{lambeta2020digit} are priced around \$400, significantly lower than the ATI Nano17 and Robotiq force torque sensors, which can exceed \$5,000. Similarly, our proposed RoTip sensor is budget-friendly \$100, and provides accurate contact surface estimation, as shown in Fig.~\ref{fig: 9}.

However, this does not imply that tactile-based adjustment is superior to force-based adjustment approaches in all aspects. Vision-based tactile sensors can experience delays due to the time needed to process extensive visual data. Although recent innovations have introduced event-based tactile sensors~\cite{funk2024evetac} with higher processing speeds, these sensors are considerably more expensive due to the costs associated with integrating delicate event-based cameras. Consequently, force sensors and force-based adjustment approaches may offer better performance for tasks that demand high-frequency responses.

\section{Conclusion} \label{Section:7}
In this paper, we propose RoTipBot, a novel approach for handling thin and flexible objects using rotatable tactile sensors. 
We first develop RoTip, a tactile sensor capable of providing comprehensive contact information around the fingertip and actively rotating its body. RoTipBot leverages RoTip's rotation capability to feed multiple sheets into the centre between the two fingers, allowing them to be grasped simultaneously. Additionally, RoTip's sensing capability is used to ensure both fingers maintain firm contact with the object and count the number of fed layers.
The results show that our RoTip sensor achieves plane estimation with a low average error of 1.51$^\circ$. Moreover, our RoTipBot outperforms state-of-the-art methods in overall success rate and operates up to three times faster. The success of RoTipBot opens new research directions for other object manipulation tasks using mobilised tactile sensors.

In the future, we aim to miniaturise the RoTip sensor and integrate it into a dexterous hand to enhance its functionality and application potential. This advancement would enable more precise and versatile manipulation of a wide range of objects. Additionally, we plan to explore the use of reinforcement learning approaches to develop a general handling policy for different kinds of deformable objects, initially guided by Finite Element Analysis (FEA). This will equip the system with the ability to autonomously learn and adapt its manipulation strategies across a variety of scenarios.
To further improve the applicability and accuracy of our approach in real-world situations, we will explore more general vision algorithms. For example, a deep learning-based corner detection model can be integrated to replace the current ArUco markers.

\appendices

\section{Supplementary for Contact Detection} 
\label{appendix:a}

For data collection, we employed a robotic arm equipped with RoTip sensors to interact with two different kinds of materials to represent thin and flexible objects, i.e., printer paper with a smooth surface and fabrics with varied textures, at various angles. A total of 392 tactile images with annotated Ground Truth (GT) masks were initially gathered from these materials, with the data collection setup illustrated in Fig.~\ref{fig: 10}. To enhance the diversity of the dataset, additional augmentation techniques such as flipping, rotation, cropping, and contrast adjustments were applied, expanding the dataset to comprise 4,000 tactile images. Then this dataset was split into 70:30 for training and testing. Key hyperparameters for the model include an SGD optimiser with a learning rate of 1e-5, momentum of 0.9, and weight decay of 5e-4, and a vanilla cross-entropy loss was used for training.

Dilated Residual Networks (DRN)~\cite{yu2017dilated} and Residual Networks (ResNet)~\cite{he2016deep} architectures were evaluated as backbones at three different input resolutions: 512, 256, and 128. The performance of each network configuration was measured using two metrics: mIoU (mean Intersection over Union) for the accuracy of segmentation tasks, and frequency representing the processing speed of the network (frames per second). All the models were tested with an Nvidia RTX-3090 graphic card for a fair comparison.  
As shown in Table~\ref{tab:segmentation}, there is a trade-off between accuracy and speed. Generally, as the input size decreases, the frequency generally increases, which suggests faster processing but potentially at the cost of reduced segmentation accuracy. However, ResNet-50 with an input size of 256 achieves a good balance between the two, with a high mIoU of 96.92\% as well as a high frequency of 173.74 Hz. Therefore, we selected ResNet-50 with a $256\times256$ input as the backbone of the network.

\begin{table}[t]
	\centering
		\caption{\footnotesize Network architecture for tactile contact segmentation.}
		\label{tab:segmentation}
        \scalebox{1}{
		\begin{tabular}{c| c | c | c}
			\hline
			Backbone & Input Size & mIoU $\uparrow$ (\%) & Frequency $\uparrow$ (Hz)   \\
			\hline
			\hline
% 			{=|=|=|=|=|=|=}
    		DRN     & 512 & 96.51 & 44.37\\
    		DRN     & 256 & 95.61 & 128.16\\
            DRN     & 128 & 90.70  & \textbf{223.81}\\
            \hline
    		ResNet-101      & 512  & 96.69 & 64.62\\
    		ResNet-101  & 256 &  96.80 & 120.28\\
    		ResNet-101 & 128 &  95.00 & 139.48\\
            \hline
            ResNet-50  & 512 &  \underline{96.83}& 79.92\\
    		ResNet-50  & 256 &  \textbf{96.92}& 173.74\\
    		ResNet-50 & 128 &   95.85& \underline{220.84}\\
    		\hline
		\end{tabular}}
\end{table}

\section{Grasp Location Analysis} \label{Section:grasp}
Unlike rigid objects, which can often be grasped at their centres~\cite{zhang2023genpose} or top surfaces~\cite{asif2018ensemblenet}, theoretical studies have yet to determine the optimal grasping positions for thin and flexible objects due to their minimal thickness and complex kinematics. Therefore, it is crucial to identify the optimal locations for squeezing and feeding these objects, prioritising two key factors: maximising handling speed and minimising the required force.

\noindent \textbf{Maximising handling speed}: 
Handling speed is directly influenced by the speed at which thin and flexible objects can be fed. Given that the RoTip sensor rotates at a constant speed, a shorter path will result in faster feeding. Therefore, it is evident that squeezing from the corners significantly reduces the movement distance of the fingers compared to squeezing from the middle or other positions, resulting in quicker handling.

\begin{figure}[t]
    \centering
    \includegraphics[width=\linewidth]{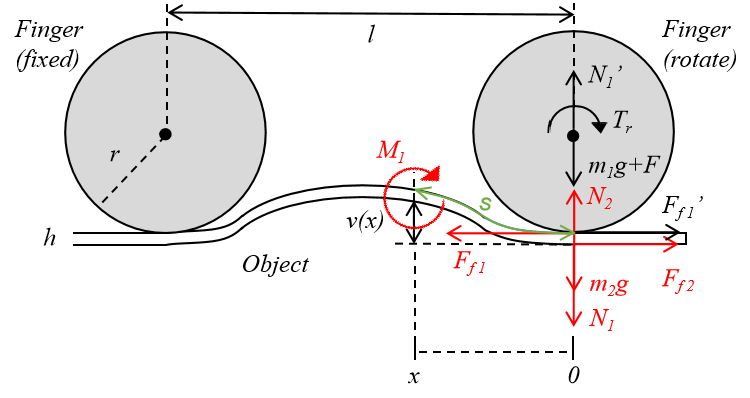}
    \caption{Theoretical force analysis of squeezing the thin and flexible object with our RoTip sensors (cross section view). The black arrows and red arrows represent the forces and torques applied to the finger and the object, respectively.}
    \label{fig: 18}
\end{figure}

\noindent \textbf{Minimising force required for handling}: It is critical to minimise the force during handling thin and flexible objects to prevent potential damage to the object while securing manipulation. We employ beam theory~\cite{audoly2008elasticity} to analyse the force involved in the deformation of such objects when they are manipulated by our proposed RoTip sensors. This section details the theoretical force analysis when the object is squeezed into a curved shape, as illustrated in Fig.~\ref{fig: 18}. 

To prevent relative sliding between the finger and the object, according to Coulomb's law and as shown in Fig.~\ref{fig: 18} for the rotating finger on the right, the friction between the finger and the object $F_{f1}{}'$ must be less than the product of static friction coefficient $\mu_{s1}$ and the normal forces generated between the object and the finger $N_{1}{}'$, i.e.,

\begin{equation}
F_{f1}{}'\leq \mu_{s1}N_{1}{}'
\label{eq:bent2}
\end{equation}

\noindent Also, when the finger maintains contact with the object, the sum of the forces acting on the fingers must equal zero, i.e.,

\begin{equation}
  \begin{cases}
    N_{1}{}'=m_{1}g+F \\
    F_{f1}{}'={T_{r}}/{r}
  \end{cases}
\end{equation}
\noindent where $m_{1}g$, $F$, $T_{r}$ and $r$ are the gravity, the vertical force, the torque applied to the finger and the finger radius, respectively.

On the other hand, to ensure that the object can continue to bend and arch, the friction between the finger and the object $F_{f1}$ applied to the object must be greater than the sum of the friction between the object and the table $F_{f2}$ and the force resulting from the rotation of the finger, i.e.,
\begin{equation}
F_{f1}\geq F_{f2}+ \frac{M_{1}}{v(x)}
\label{eq:bent1}
\end{equation}
\noindent where $M_{1}$ and $v(x)$ represent the bending moment and the vertical deflection at position $x$, respectively; From Fig.~\ref{fig: 18}, we know that:
\begin{equation}
    F_{f2}=\mu _{k2}N_{2}=\mu _{k2}(m_{2}g+N_{1})=\mu _{k2}(m_{2}g+m_{1}g+F)
\end{equation}
\noindent where $\mu _{k2}$ is the dynamic friction coefficient of $F_{f2}$ and $m_{2}g$ is the gravity of the object.

\begin{figure}
    \centering
    \includegraphics[width=\linewidth]{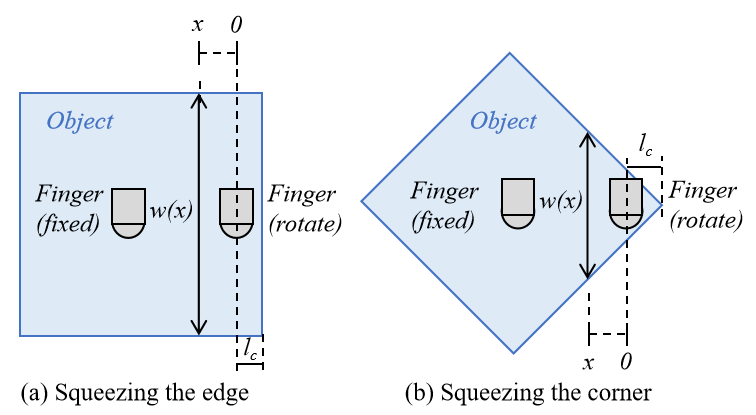}
    \caption{Finger squeezing at (a) the edge and (b) the corner of the object.}
    \label{fig: 19}
\end{figure}

According to beam theory~\cite{audoly2008elasticity}, there are two cases for calculating $M_{1}$: small deflection and large deflection. These cases are defined based on whether the deflection $v(x)$ reaches one-tenth of the distance $l$ between the two fingers:

\begin{equation}
  M_{1}=
  \begin{cases}
    EI_{z}v{}''(x), & \text{when } v(x) \leq \frac{l}{10}\\
    EI_{z}\frac{ds}{d\kappa }, & \text{when } v(x) > \frac{l}{10}
  \end{cases}
  \label{eq:RotationMoment}
\end{equation}

\noindent where $E$ is the elastic modulus, $v{}''(x)$ is the curvature of the object (the second derivative of $v(x)$), $s$ is the arc length along the deﬂected object between position $0$ end and position $x$ as shown in Fig.~\ref{fig: 18}, and $\kappa $ is the slope at position $x$; $I_{z}$ is the moment of inertia of the object:

\begin{equation}
    I_{z}=\int_{0}^{l} \frac{w(x)h^{3}}{12} dx
    \label{eq:InertiaMoment}
\end{equation} 
\noindent where $w(x)$ is the width of the contact edge at position $x$, as shown in Fig.~\ref{fig: 19}, and $h$ is the thickness of each layer.  

From Eqs.~\ref{eq:bent1}-\ref{eq:InertiaMoment}, we can find that the minimum force required to squeeze the thin and flexible object has a positive correlation with the ratio $\frac{M_{1}}{v(x)}$; Considering the same bending curve of the object at different squeeze locations, the deflection $v(x)$ and the curvature $v{}''(x)$ or $\frac{ds}{d\kappa }$ of the object remain the same. Hence, the optimal location for grasping is where $M_{1}$ is minimised, which occurs at the corners of the object due to the minimal $\int_{0}^{l} w(x)dx$ based on Eq.~\ref{eq:RotationMoment}-\ref{eq:InertiaMoment}, and as shown in Fig.~\ref{fig: 19}. Therefore, squeezing at the corners of the object results in minimal force for effective handling.

\section{Configuration of FEM Simulation}
\label{appendix:b}

\textbf{Assembling Setup}:
The assembling model consists of three parts: a \textit{30 cm $\times$ 30 cm} shell plane set as the deformable object, a \textit{40 cm $\times$ 40 cm $\times$ 4 cm} rigid solid set as the tabletop, and two soft finger membranes. The target object is placed at the centre of the tabletop, with the sensors placed parallel to each other at a spacing of 80 cm, pressing onto the centre, edge, and corner of the paper, respectively.

\textbf{Property}:
Here, we consider the deformable object as an office paper to simulate the grasping process. The material parameters of the paper based on the elastic model were set to \textit{Mass Density=1.2e-9 ton/mm$^3$}, \textit{Young's Modulus=828 MPa}, \textit{Poisson's Ratio=0.3}~\cite{lee2016bending}. The finger membrane was modelled using the Neo-Hookean hyperelastic model, with \textit{Mass Density=1e-9 ton/mm$^3$}, \textit{C$_{10}$=0.0335}, \textit{D$_1$=1.2297}~\cite{insitu2023}.

\textbf{Step}: 
To replicate the actual dynamic contact process between the fingers and the paper, we chose the \textit{Dynamic Explicit Analysis} method and divided the interaction into two steps. In the first step, the finger presses the paper; in the second step, the finger grasps the paper through rotation, with each step lasting one second. The output measured is the total force due to the frictional stress between the fingers and the paper, allowing us to compare the magnitude of the frictional forces required at different contact locations.

\textbf{Interaction}:
We have defined two types of contacts: the contact between the fingers and the paper; and the contact between the paper and the tabletop. For both types, normal contacts and tangential contacts are defined. Tangential contact is defined using a penalty friction formulation, with the friction coefficients set as 0.5 and 0.1, respectively~\cite{skedung2011tactile}.

\textbf{Load}:
The fingers will first execute a translational movement at a speed of 0.4 mm/s, pressing the paper in the direction perpendicular to the tabletop until the displacement reaches 0.4 mm. Then, the two fingers will rotate in opposite directions at a speed of 2 rad/s, twisting the paper from the table until the rotation angle reaches 2 radians. The tabletop remains fixed throughout the entire process.  

\section{RoTip Calibration}
To achieve accurate tactile perception with RoTip sensors, two key components need to be calibrated: the camera intrinsic matrix, and the extrinsic offset $(o_x, o_y, o_z)$ as shown in Fig.~\ref{fig: 5}. Similar to the method in~\cite{yuan2023tactile}, we use a 7x8 checkerboard mounted on a board to get the sensor intrinsic with ROS calibration tools. During camera calibration, multiple sensor images are collected with different checkerboard poses to compute the camera intrinsic matrix $K$. Next, we calibrate extrinsic offsets $o_x$ and $o_y$ using tactile images from vertical contact with a flat surface, ensuring the contact area is centred on the fingertip. This step is treated as an optimisation problem aiming to minimise the distance between the predicted contact region's centre and the fingertip's top.
Finally, we calibrate the offset $o_z$ along the camera's $z$-axis by minimising the error between the predicted contact plane and the actual contact plane, using the tactile images collected at tilted angles, specifically 45$^\circ$ and -45$^\circ$. 
%\noindent \textbf{Remark:} 
We chose not to adopt an end-to-end approach for contact surface prediction because it would require training a model for each individual RoTip sensor. Instead, by calibrating only the offset parameters $(o_x, o_y, o_z)$ for each sensor, our approach simplifies the calibration process and enhances the model's generalisation across RoTip sensors with varying installation errors.

\bibliographystyle{IEEEtran}
\bibliography{egibib}

\begin{thebibliography}{10}
\providecommand{\url}[1]{#1}
\csname url@samestyle\endcsname
\providecommand{\newblock}{\relax}
\providecommand{\bibinfo}[2]{#2}
\providecommand{\BIBentrySTDinterwordspacing}{\spaceskip=0pt\relax}
\providecommand{\BIBentryALTinterwordstretchfactor}{4}
\providecommand{\BIBentryALTinterwordspacing}{\spaceskip=\fontdimen2\font plus
\BIBentryALTinterwordstretchfactor\fontdimen3\font minus \fontdimen4\font\relax}
\providecommand{\BIBforeignlanguage}[2]{{%
\expandafter\ifx\csname l@#1\endcsname\relax
\typeout{** WARNING: IEEEtran.bst: No hyphenation pattern has been}%
\typeout{** loaded for the language `#1'. Using the pattern for}%
\typeout{** the default language instead.}%
\else
\language=\csname l@#1\endcsname
\fi
#2}}
\providecommand{\BIBdecl}{\relax}
\BIBdecl

\bibitem{flipbot}
C.~Zhao, C.~Jiang, J.~Cai, M.~Y. Wang, H.~Yu, and Q.~Chen, ``Flipbot: Learning continuous paper flipping via coarse-to-fine exteroceptive-proprioceptive exploration,'' in \emph{Proc. IEEE Int. Conf. Robot. Autom.}, 2023, pp. 10\,282--10\,288.

\bibitem{zhu2022challenges}
J.~Zhu, A.~Cherubini, C.~Dune, D.~Navarro-Alarcon, F.~Alambeigi, D.~Berenson, F.~Ficuciello, K.~Harada, J.~Kober, X.~Li \emph{et~al.}, ``Challenges and outlook in robotic manipulation of deformable objects,'' \emph{IEEE Robot. Automa. Mag.}, vol.~29, no.~3, pp. 67--77, 2022.

\bibitem{billard2019trends}
A.~Billard and D.~Kragic, ``Trends and challenges in robot manipulation,'' \emph{Science}, vol. 364, no. 6446, p. eaat8414, 2019.

\bibitem{teeple2022multi}
C.~B. Teeple, J.~Werfel, and R.~J. Wood, ``Multi-dimensional compliance of soft grippers enables gentle interaction with thin, flexible objects,'' in \emph{Proc. IEEE Int. Conf. Robot. Autom.}, 2022, pp. 728--734.

\bibitem{zheng2022autonomous}
Y.~Zheng, F.~F. Veiga, J.~Peters, and V.~J. Santos, ``Autonomous learning of page flipping movements via tactile feedback,'' \emph{IEEE Trans. Robot.}, vol.~38, no.~5, pp. 2734--2749, 2022.

\bibitem{koivikko20213d}
A.~Koivikko, D.-M. Drotlef, C.~B. Dayan, V.~Sariola, and M.~Sitti, ``3d-printed pneumatically controlled soft suction cups for gripping fragile, small, and rough objects,'' \emph{Adv. Intell. Syst.}, vol.~3, no.~9, 2021.

\bibitem{chin2020multiplexed}
L.~Chin, F.~Barscevicius, J.~Lipton, and D.~Rus, ``Multiplexed manipulation: Versatile multimodal grasping via a hybrid soft gripper,'' in \emph{Proc. IEEE Int. Conf. Robot. Autom.}, 2020, pp. 8949--8955.

\bibitem{jiang2019dynamic}
C.~Jiang, A.~Nazir, G.~Abbasnejad, and J.~Seo, ``Dynamic flex-and-flip manipulation of deformable linear objects,'' in \emph{Proc. IEEE/RSJ Int. Conf. Intell. Robots Syst.}, 2019, pp. 3158--3163.

\bibitem{yuan2017gelsight}
W.~Yuan, S.~Dong, and E.~H. Adelson, ``Gelsight: High-resolution robot tactile sensors for estimating geometry and force,'' \emph{Sensors}, vol.~17, no.~12, p. 2762, 2017.

\bibitem{ward2018tactip}
B.~Ward-Cherrier, N.~Pestell, L.~Cramphorn, B.~Winstone, M.~E. Giannaccini, J.~Rossiter, and N.~F. Lepora, ``The tactip family: Soft optical tactile sensors with 3d-printed biomimetic morphologies,'' \emph{Soft Robot.}, vol.~5, no.~2, pp. 216--227, 2018.

\bibitem{johnson2009retrographic}
M.~K. Johnson and E.~H. Adelson, ``Retrographic sensing for the measurement of surface texture and shape,'' in \emph{Proc. IEEE Conf. Comput. Vis. Pattern Recognit.}, 2009, pp. 1070--1077.

\bibitem{taylor2022gelslim}
I.~H. Taylor, S.~Dong, and A.~Rodriguez, ``Gelslim 3.0: High-resolution measurement of shape, force and slip in a compact tactile-sensing finger,'' in \emph{Proc. IEEE Int. Conf. Robot. Autom.}, 2022, pp. 10\,781--10\,787.

\bibitem{lambeta2020digit}
M.~Lambeta, P.-W. Chou, S.~Tian, B.~Yang, B.~Maloon, V.~R. Most, D.~Stroud, R.~Santos, A.~Byagowi, G.~Kammerer \emph{et~al.}, ``Digit: A novel design for a low-cost compact high-resolution tactile sensor with application to in-hand manipulation,'' \emph{IEEE Robot. Autom. Lett.}, vol.~5, no.~3, pp. 3838--3845, 2020.

\bibitem{wang2021gelsight}
S.~Wang, Y.~She, B.~Romero, and E.~Adelson, ``Gelsight wedge: Measuring high-resolution 3d contact geometry with a compact robot finger,'' in \emph{Proc. IEEE Int. Conf. Robot. Autom.}, 2021, pp. 6468--6475.

\bibitem{gomes2020geltip}
D.~F. Gomes, Z.~Lin, and S.~Luo, ``Geltip: A finger-shaped optical tactile sensor for robotic manipulation,'' in \emph{Proc. IEEE/RSJ Int. Conf. Intell. Robots Syst.}, 2020, pp. 9903--9909.

\bibitem{gomes2020blocks}
------, ``Blocks world of touch: Exploiting the advantages of all-around finger sensing in robot grasping,'' \emph{Front. Robot. AI}, vol.~7, p. 127, 2020.

\bibitem{gomes2023beyond}
D.~F. Gomes, P.~Paoletti, and S.~Luo, ``{Beyond Flat GelSight Sensors: Simulation of Optical Tactile Sensors of Complex Morphologies for Sim2Real Learning},'' in \emph{Proc. RSS}, 2023.

\bibitem{padmanabha2020omnitact}
A.~Padmanabha, F.~Ebert, S.~Tian, R.~Calandra, C.~Finn, and S.~Levine, ``Omnitact: A multi-directional high-resolution touch sensor,'' in \emph{Proc. IEEE Int. Conf. Robot. Autom.}, 2020, pp. 618--624.

\bibitem{sun2022soft}
H.~Sun, K.~J. Kuchenbecker, and G.~Martius, ``A soft thumb-sized vision-based sensor with accurate all-round force perception,'' \emph{Nat. Mach. Intell.}, vol.~4, no.~2, pp. 135--145, 2022.

\bibitem{tippur2023gelsight360}
M.~H. Tippur and E.~H. Adelson, ``Gelsight360: An omnidirectional camera-based tactile sensor for dexterous robotic manipulation,'' in \emph{Proc. IEEE Int. Conf. Soft Robot.}, 2023, pp. 1--8.

\bibitem{tacrot}
W.~Zhang, C.~Xia, X.~Zhu, H.~Liu, and B.~Liang, ``Tacrot: A parallel-jaw gripper with rotatable tactile sensors for in-hand manipulation,'' in \emph{Proc. IEEE Int. Conf. Syst. Man. Cybern.}, 2022, pp. 423--429.

\bibitem{yuan2023tactile}
S.~Yuan, S.~Wang, R.~Patel, M.~Tippur, C.~L. Yako, M.~R. Cutkosky, E.~Adelson, and J.~K. Salisbury, ``Tactile-reactive roller grasper,'' \emph{IEEE Trans. Robot.}, vol.~41, pp. 1938--1955, 2025.

\bibitem{cao2023touchroller}
G.~Cao, J.~Jiang, C.~Lu, D.~F. Gomes, and S.~Luo, ``Touchroller: A rolling optical tactile sensor for rapid assessment of textures for large surface areas,'' \emph{Sensors}, vol.~23, no.~5, p. 2661, 2023.

\bibitem{pan2023hand}
C.~Pan, M.~Lepert, S.~Yuan, R.~Antonova, and J.~Bohg, ``In-hand manipulation of unknown objects with tactile sensing for insertion,'' in \emph{Proc. IEEE/RSJ Int. Conf. Intell. Robots Syst.}, 2023, pp. 8765--8771.

\bibitem{jin2019robust}
S.~Jin, C.~Wang, and M.~Tomizuka, ``Robust deformation model approximation for robotic cable manipulation,'' in \emph{Proc. IEEE/RSJ Int. Conf. Intell. Robots Syst.}, 2019, pp. 6586--6593.

\bibitem{she2021cable}
Y.~She, S.~Wang, S.~Dong, N.~Sunil, A.~Rodriguez, and E.~Adelson, ``Cable manipulation with a tactile-reactive gripper,'' \emph{Int. J. Robot. Res.}, vol.~40, no. 12-14, pp. 1385--1401, 2021.

\bibitem{pecyna2022visual}
L.~Pecyna, S.~Dong, and S.~Luo, ``Visual-tactile multimodality for following deformable linear objects using reinforcement learning,'' in \emph{Proc. IEEE/RSJ Int. Conf. Intell. Robots Syst.}, 2022, pp. 3987--3994.

\bibitem{zhang2019probabilistic}
F.~Zhang, A.~Cully, and Y.~Demiris, ``Probabilistic real-time user posture tracking for personalized robot-assisted dressing,'' \emph{IEEE Trans. Robot.}, vol.~35, no.~4, pp. 873--888, 2019.

\bibitem{lee2021learning}
R.~Lee, D.~Ward, V.~Dasagi, A.~Cosgun, J.~Leitner, and P.~Corke, ``Learning arbitrary-goal fabric folding with one hour of real robot experience,'' in \emph{Proc. Conf. Robot Learn.}, 2021, pp. 2317--2327.

\bibitem{sunil2023visuotactile}
N.~Sunil, S.~Wang, Y.~She, E.~Adelson, and A.~R. Garcia, ``Visuotactile affordances for cloth manipulation with local control,'' in \emph{Proc. Conf. Robot Learn.}, 2023, pp. 1596--1606.

\bibitem{2023arXiv230102749Z}
J.~Zhu, M.~Gienger, G.~Franzese, and J.~Kober, ``Do you need a hand? – a bimanual robotic dressing assistance scheme,'' \emph{IEEE Trans. Robot.}, vol.~40, pp. 1906--1919, 2024.

\bibitem{navarro2017fourier}
D.~Navarro-Alarcon and Y.-H. Liu, ``Fourier-based shape servoing: A new feedback method to actively deform soft objects into desired 2-d image contours,'' \emph{IEEE Trans. Robot.}, vol.~34, no.~1, pp. 272--279, 2017.

\bibitem{li2022contact}
S.~Li, Z.~Huang, T.~Du, H.~Su, J.~B. Tenenbaum, and C.~Gan, ``Contact points discovery for soft-body manipulations with differentiable physics,'' in \emph{Proc. Int. Conf. Learn. Representations}, 2022.

\bibitem{zhu2019robotic}
J.~Zhu, B.~Navarro, R.~Passama, P.~Fraisse, A.~Crosnier, and A.~Cherubini, ``Robotic manipulation planning for shaping deformable linear objects withenvironmental contacts,'' \emph{IEEE Robot. Autom. Lett.}, vol.~5, no.~1, pp. 16--23, 2019.

\bibitem{qiu2023robotic}
Y.~Qiu, J.~Zhu, C.~Della~Santina, M.~Gienger, and J.~Kober, ``Robotic fabric flattening with wrinkle direction detection,'' in \emph{Proc. Int. Symp. Exp. Robot.}, 2023, pp. 339--350.

\bibitem{hang2019pre}
K.~Hang, A.~S. Morgan, and A.~M. Dollar, ``Pre-grasp sliding manipulation of thin objects using soft, compliant, or underactuated hands,'' \emph{IEEE Robot. Autom. Lett.}, vol.~4, no.~2, pp. 662--669, 2019.

\bibitem{babin2019stable}
V.~Babin, D.~St-Onge, and C.~Gosselin, ``Stable and repeatable grasping of flat objects on hard surfaces using passive and epicyclic mechanisms,'' \emph{Robot. Comput.-Integr. Manuf.}, vol.~55, pp. 1--10, 2019.

\bibitem{zhang2022prying}
Q.~Zhang, Z.~Hu, K.~Koyama, W.~Wan, and K.~Harada, ``Prying grasp for picking thin object using thick fingertips,'' \emph{IEEE Robot. Autom. Lett.}, vol.~7, no.~4, pp. 11\,577--11\,584, 2022.

\bibitem{babin2018picking}
V.~Babin and C.~Gosselin, ``Picking, grasping, or scooping small objects lying on flat surfaces: A design approach,'' \emph{Int. J. Robot. Res.}, vol.~37, no.~12, pp. 1484--1499, 2018.

\bibitem{ko2020tendon}
T.~Ko, ``A tendon-driven robot gripper with passively switchable underactuated surface and its physics simulation based parameter optimization,'' \emph{IEEE Robot. Autom. Lett.}, vol.~5, no.~4, pp. 5002--5009, 2020.

\bibitem{agboh2022multi}
W.~C. Agboh, J.~Ichnowski, K.~Goldberg, and M.~R. Dogar, ``Multi-object grasping in the plane,'' in \emph{Int. Symp. Robot. Res.}\hskip 1em plus 0.5em minus 0.4em\relax Springer, 2022, pp. 222--238.

\bibitem{li2024grasp}
Y.~Li, B.~Liu, Y.~Geng, P.~Li, Y.~Yang, Y.~Zhu, T.~Liu, and S.~Huang, ``Grasp multiple objects with one hand,'' \emph{IEEE Robot. Auto. Lett.}, 2024.

\bibitem{chen2018encoder}
L.-C. Chen, Y.~Zhu, G.~Papandreou, F.~Schroff, and H.~Adam, ``Encoder-decoder with atrous separable convolution for semantic image segmentation,'' in \emph{Proc. Eur. Conf. Comput. Vis.}, 2018, pp. 801--818.

\bibitem{szeliski2022computer}
R.~Szeliski, \emph{Computer vision: algorithms and applications}.\hskip 1em plus 0.5em minus 0.4em\relax Springer Nature, 2022.

\bibitem{azulay2023allsight}
O.~Azulay, N.~Curtis, R.~Sokolovsky, G.~Levitski, D.~Slomovik, G.~Lilling, and A.~Sintov, ``Allsight: A low-cost and high-resolution round tactile sensor with zero-shot learning capability,'' \emph{IEEE Robot. Autom. Lett.}, vol.~9, no.~1, pp. 483--490, 2023.

\bibitem{dzyaloshinskii1961general}
I.~E. Dzyaloshinskii, E.~M. Lifshitz, and L.~P. Pitaevskii, ``The general theory of van der waals forces,'' \emph{Adv. Phys.}, vol.~10, no.~38, pp. 165--209, 1961.

\bibitem{lin2023gelfinger}
Z.~Lin, J.~Zhuang, Y.~Li, X.~Wu, S.~Luo, D.~F. Gomes, F.~Huang, and Z.~Yang, ``Gelfinger: A novel visual-tactile sensor with multi-angle tactile image stitching,'' \emph{IEEE Robot. Autom. Lett.}, 2023.

\bibitem{shimonomura2021detection}
K.~Shimonomura, T.~Chang, and T.~Murata, ``Detection of foreign bodies in soft foods employing tactile image sensor,'' \emph{Front. Robot. AI.}, vol.~8, p. 774080, 2021.

\bibitem{xu2024dtactive}
J.~Xu, L.~Wu, C.~Lin, D.~Zhao, and H.~Xu, ``Dtactive: A vision-based tactile sensor with active surface,'' \emph{arXiv preprint arXiv:2410.08337}, 2024.

\bibitem{calandra2018more}
R.~Calandra, A.~Owens, D.~Jayaraman, J.~Lin, W.~Yuan, J.~Malik, E.~H. Adelson, and S.~Levine, ``More than a feeling: Learning to grasp and regrasp using vision and touch,'' \emph{IEEE Robot. Autom. Lett.}, vol.~3, no.~4, pp. 3300--3307, 2018.

\bibitem{liang2023visuo}
W.~Liang, F.~Fang, C.~Acar, W.~Q. Toh, Y.~Sun, Q.~Xu, and Y.~Wu, ``Visuo-tactile feedback-based robot manipulation for object packing,'' \emph{IEEE Robot. Autom. Lett.}, vol.~8, no.~2, pp. 1151--1158, 2023.

\bibitem{hogan2018tactile}
F.~R. Hogan, M.~Bauza, O.~Canal, E.~Donlon, and A.~Rodriguez, ``Tactile regrasp: Grasp adjustments via simulated tactile transformations,'' in \emph{Proc. IEEE/RSJ Int. Conf. Intell. Robots Syst.}, 2018, pp. 2963--2970.

\bibitem{2022jiaqi}
J.~Jiang, G.~Cao, A.~Butterworth, T.-T. Do, and S.~Luo, ``Where shall i touch? vision-guided tactile poking for transparent object grasping,'' \emph{IEEE/ASME Trans. Mechatron.}, vol.~28, no.~1, pp. 233--244, 2022.

\bibitem{lloyd2023pose}
J.~Lloyd and N.~F. Lepora, ``Pose-and-shear-based tactile servoing,'' \emph{Int. J. Robot. Res.}, vol.~43, no.~7, pp. 1024--1055, 2024.

\bibitem{funk2024evetac}
N.~Funk, E.~Helmut, G.~Chalvatzaki, R.~Calandra, and J.~Peters, ``Evetac: An event-based optical tactile sensor for robotic manipulation,'' \emph{IEEE Trans. Robot.}, 2024.

\bibitem{yu2017dilated}
F.~Yu, V.~Koltun, and T.~Funkhouser, ``Dilated residual networks,'' in \emph{Proc. IEEE Conf. Comput. Vis. Pattern Recognit.}, 2017, pp. 472--480.

\bibitem{he2016deep}
K.~He, X.~Zhang, S.~Ren, and J.~Sun, ``Deep residual learning for image recognition,'' in \emph{Proc. IEEE Conf. Comput. Vis. Pattern Recognit.}, 2016, pp. 770--778.

\bibitem{zhang2023genpose}
J.~Zhang, M.~Wu, and H.~Dong, ``Genpose: Generative category-level object pose estimation via diffusion models,'' \emph{Adv. Neural Inf. Process. Syst.}, 2023.

\bibitem{asif2018ensemblenet}
U.~Asif, J.~Tang, and S.~Harrer, ``Ensemblenet: Improving grasp detection using an ensemble of convolutional neural networks.'' in \emph{BMVC}, 2018.

\bibitem{audoly2008elasticity}
B.~Audoly and Y.~Pomeau, \emph{Elasticity and geometry: from hair curls to the non-linear response of shells}.\hskip 1em plus 0.5em minus 0.4em\relax Oxford University Press, 2008.

\bibitem{lee2016bending}
M.~Lee, S.~Kim, H.-Y. Kim, and L.~Mahadevan, ``Bending and buckling of wet paper,'' \emph{Phys. Fluids}, vol.~28, no.~4, 2016.

\bibitem{insitu2023}
C.~Zhao, J.~Ren, H.~Yu, and D.~Ma, ``In-situ mechanical calibration for vision-based tactile sensors,'' in \emph{Proc. IEEE Int. Conf. Robot. Autom.}, 2023, pp. 10\,387--10\,393.

\bibitem{skedung2011tactile}
L.~Skedung, K.~Danerl{\"o}v, U.~Olofsson, C.~M. Johannesson, M.~Aikala, J.~Kettle, M.~Arvidsson, B.~Berglund, and M.~W. Rutland, ``Tactile perception: Finger friction, surface roughness and perceived coarseness,'' \emph{Tribol. Int.}, vol.~44, no.~5, pp. 505--512, 2011.

\end{thebibliography}

% \subsection{xxx} 

\vfill

\end{document}